\documentclass[twoside,11pt]{article}

\usepackage{jmlr2e}

\usepackage{amsmath,bm,url}
\usepackage{amsfonts}
\usepackage{multirow}
\usepackage{color}
\usepackage{subfigure} 
\usepackage{algorithm}
\usepackage{algorithmic}

\def\Ic{{\cal I}}

\def\qed{\hfill\hbox{${\vcenter{\vbox{
    \hrule height 0.4pt\hbox{\vrule width 0.4pt height 6pt
    \kern5pt\vrule width 0.4pt}\hrule height 0.4pt}}}$}}
    
\def\Depth{\delta}

\usepackage{multirow}

\newcommand{\rc}{r}

\newcommand{\RG}{\mathrm{RG}}
\newcommand{\EC}{\mathrm{EC}}

\newcommand{\TestNodeRelationships}{\mathsf{TestNodeRelationships}}

\newcommand{\calP}{\mathcal{P}}
\newcommand{\calJ}{\mathcal{J}}

\newcommand{\calE}{\mathcal{E}}

\newcommand{\calX}{\mathcal{X}}
\newcommand{\calA}{\mathcal{A}}

\newcommand{\bX}{\mathbf{X}}

\newcommand{\bx}{\mathbf{x}}
\newcommand{\bd}{\mathbf{d}}

\newcommand{\hp}{\widehat{p}}
\newcommand{\hd}{\widehat{d}}

\newcommand{\hbD}{\mathbf{\widehat{D}}}
\newcommand{\hPhi}{\widehat{\Phi}}

\newcommand{\calC}{\mathcal{C}}

\newcommand{\calT}{\mathcal{T}}
\newcommand{\bD}{\mathbf{D}}
\newcommand{\calK}{\mathcal{K}}
\newcommand{\bN}{\mathbb{N}}

\newcommand{\hT}{\widehat{T}}
\newcommand{\ML}{\mathrm{CL}}
\newcommand{\Sg}{\mathrm{Sg}}
\newcommand{\old}{\mathrm{old}}
\newcommand{\new}{\mathrm{new}}

\newcommand{\Path}{\mathrm{Path}}
\newcommand{\CL}{\mathrm{CL}}
\newcommand{\blind}{\mathrm{blind}}
\newcommand{\Blind}{\mathrm{BlindTransform}}
\newcommand{\diam}{\mathrm{diam}}
\newcommand{\MST}{\mathrm{MST}}
\newcommand{\Leaf}{\mathrm{Leaf}}

\newcommand{\hrho}{\widehat{\rho}}
\newcommand{\htheta}{\widehat{\theta}}
\newcommand{\bSigma}{\bm{\Sigma}}
\newcommand{\hSigma}{\widehat{\Sigma}}

\newcommand{\nbd}{\mathrm{nbd}}
\newcommand{\bR}{\mathbb{R}}

\newcommand{\hdiss}{\widehat{\Lambda}}
\newcommand{\hbdiss}{\widehat{\mathbf{\Lambda}}}

\DeclareMathOperator*{\argmax}{argmax}
\DeclareMathOperator*{\argmin}{argmin}

\newtheorem{THEO}{\bf Theorem}

\newtheorem{LEMM}[THEO]{\bf Lemma}

\newtheorem{DEFI}[THEO]{\bf Definition}

\newtheorem{PROP}[THEO]{\bf Proposition}

\newcommand{\bprf}{\begin{myproof}}
\newcommand{\eprf}{\end{myproof}}

\newenvironment{proof_of}[1]{\noindent {\em Proof of #1: }}{\hspace*{\fill} $\Box$ }
\newenvironment{myproof}{\noindent{\em Proof:} \hspace*{1em}}{
    \hspace*{\fill} $\Box$ }

\ShortHeadings{Learning Latent Tree Graphical Models}{Choi, Tan, Anandkumar, and Willsky}

\begin{document} 

\title{Learning Latent Tree Graphical Models}

\author{\name Myung Jin Choi{$^\dagger$} \email myungjin@mit.edu \\
\name Vincent Y. F. Tan{$^\dagger$} \email vtan@mit.edu \\
\name Animashree Anandkumar{$^\ddagger$} \email a.anandkumar@uci.edu \\
\name Alan S. Willsky{$^\dagger$} \email willsky@mit.edu \\
\addr $^\dagger$ Stochastic Systems Group,\\
Laboratory for Information and Decision Systems,\\
Massachusetts Institute of Technology. \\
$^\ddagger$ Center for Pervasive Communications and Computing,\\
Electrical Engineering and Computer Science, \\
University of California, Irvine. }
\editor{}

\maketitle

\begin{abstract}%
We study the problem of learning a latent tree graphical model where samples are available only from a subset of variables. We propose two consistent and computationally efficient algorithms for learning {\em minimal} latent trees, that is, trees without any redundant hidden nodes.  Unlike many existing methods, the observed nodes (or variables) are not constrained to be leaf nodes. Our first algorithm, {\em recursive grouping}, builds the latent tree recursively by identifying sibling groups using   so-called information distances. One of the main contributions of this work is our second algorithm, which we refer to as {\em CLGrouping}. CLGrouping starts with a pre-processing procedure in which a tree over the observed variables is constructed. This global step groups the observed nodes that are likely to be close to each other in the true latent tree, thereby guiding subsequent recursive grouping (or   equivalent procedures) on much smaller subsets of variables. This results in more accurate and efficient learning of latent trees. We also present regularized versions of our algorithms that learn latent tree approximations of arbitrary distributions. We compare the proposed algorithms to other methods by  performing extensive numerical experiments on various latent tree graphical models such as hidden Markov models and star graphs.  In addition, we demonstrate the applicability of our methods on real-world datasets by modeling the dependency structure of monthly stock returns in the S\&P index and of the words in the 20 newsgroups dataset. 
\end{abstract}

\begin{keywords}
  Graphical Models, Hidden Variables, Latent Tree Models, Structure Learning
\end{keywords}

\section{Introduction} \label{sec:intro}

The inclusion of latent variables in modeling complex phenomena and data is a well-recognized and a valuable construct in a variety of applications, including bioinformatics and computer vision, and the investigation of machine-learning methods for models with latent variables is a substantial and continuing direction of research.
  
There are three challenging problems in learning a model with latent variables: learning the number of latent variables; inferring the structure of how these latent variables relate to each other and to the observed variables; and estimating the parameters characterizing those relationships.  Issues that one must consider in developing a new learning algorithm include developing tractable methods; incorporating the tradeoff between the fidelity to the given data and generalizability; deriving theoretical results on the performance of such algorithms;  and studying applications that provide clear motivation and contexts for the models so learned.

One class of models that has received considerable attention in the literature is the class of {\em latent tree models}, i.e., graphical models Markov on trees, in which variables at some nodes represent the  original (observed) variables of interest while others represent  the latent variables.  The appeal of such models for computational tractability is clear: with a tree-structured model  describing the statistical relationships, inference - processing noisy observations of some or all of the original variables to compute the estimates of all variables - is straightforward and scalable.  Although the class of tree-structured models, with or without latent variables, is a constrained one, there are interesting applications that provide strong motivation for the work presented here.  In particular, a very active avenue of research in computer vision is the use of context - e.g., the nature of a scene to aid the reliable recognition of objects (and at the same time to allow the recognition of particular objects to assist in recognizing the scene). For example, if one knows that an image is that of an office, then one might expect to find a desk, a monitor on that desk, and perhaps a computer mouse.  Hence if one builds a model with a latent variable representing that context (``office'') and uses simple, noisy detectors for different object types, one would expect that the detection of a desk would support the likelihood that one is looking at an office and through that enhance the reliability of detecting smaller objects (monitors, keyboards, mice, etc.).  Work along these lines, including by some of the authors of this paper \citep{parikh,choi10}, show the promise of using tree-based models of context.

This paper considers the problem of learning tree-structured latent models.  If all variables are observed in the tree under consideration, then the well-known algorithm of \cite{chow68} provides a tractable algorithm for performing maximum likelihood (ML) estimation of the tree structure.  However, ML estimation of latent tree models is NP-hard \citep{Roch:CompBio}.  This has motivated a number of investigations of other tractable methods for learning such trees as well as theoretical guarantees on performance.  Our work represents a contribution to this area of investigation.

There are three main contributions in our paper.  Firstly, by adopting a statistical distance-based framework, we develop two new algorithms, recursive grouping and CLGrouping, for the learning of latent trees, which applies equally well to discrete and Gaussian models. Secondly, we provide consistency guarantees (both structural and parametric) as well as very favorable computational and sample complexity characterizations for both of our algorithms.  Thirdly, through extensive numerical experiments on both synthetic and real-world data, we demonstrate the superiority of our approach for a wide variety of models ranging from ones with very large tree diameters (e.g., hidden Markov models (HMMs)) to star models and  complete trees.\footnote{A {\em complete} $k$-ary tree (or $k$-complete tree) is one in which all leaf nodes are at same depth and all internal nodes have degree $k$. }

Our first algorithm, which we refer to as {\em recursive grouping}, constructs a latent tree in a bottom-up fashion, grouping nodes into sibling groups that share the same parent node, recursively at each level of the resulting hierarchy (and allowing for some of the observed variables to play roles at arbitrary levels in the resulting hierarchy).  Our second algorithm, {\em CLGrouping} first implements a global construction step, namely producing the Chow-Liu tree for the observed variables without any hidden nodes.  This global step then provides guidance for groups of observed nodes that are likely to be topologically close to each other in the latent tree, thereby guiding subsequent recursive grouping or neighbor-joining \citep{Sai87} computations.  Each of these algorithms is consistent and has excellent sample  and computational complexity.\footnote{As we will see, depending on the true latent tree model, one or the other of these may be more efficient.  Roughly speaking, for smaller diameter graphs (such as the star), recursive grouping is faster, and for larger diameter graphs (such as an HMM), CLgrouping is more efficient. }

As \cite{pearl88} points out, the identification of latent tree models has some built-in ambiguity, as there is an entire equivalence class of models in the sense that when all latent variables are marginalized out, each model in this class yields the same joint distribution over the observed variables.  For example, we can take any such latent model and add another hidden variable as a leaf node connected to only one other (hidden or observed) node.  Hence, much as one finds in fields such as state space dynamic systems (e.g., \citet[Section 8]{Lue79}), there is a notion of minimality that is required here, and our results are stated in terms of consistent learning of such minimal latent models.

\subsection{Related Work}\label{sec:related}
The relevant literature on learning latent models is vast and  in this section, we summarize the main lines of research in this area. 


The classical {\em latent cluster models} (LCM) consider multivariate distributions in which there exists only {\em one} latent variable and each state of the variable corresponds to a cluster in the data \citep{lazarsfeld68}.  Hierarchical latent class (HLC) models \citep{zhang04,ZhangJMLR04,Che08} generalize these models by allowing multiple latent variables. 
HLC allows latent variables to have different number of states, but assume that all observed nodes are at the leaves of the tree.  Their learning algorithm is based on a greedy approach of making one local move at a time (e.g., introducing one hidden node, or replacing an edge), which is computationally expensive and does not have consistency guarantees.  Another greedy learning algorithm called BIN \citep{harmeling10} is computationally more efficient, but enforces that each internal node is hidden and has three neighboring nodes.
In contrast, we fix the number of states in each hidden node, but allow observed nodes to be internal nodes.  Our algorithms are guaranteed to recover the correct structure when certain (mild) conditions are met.  

Many authors also propose reconstructing latent trees using the expectation  maximization (EM) algorithm \citep{elidan05, kemp08}. However, as with all other EM-based methods, these approaches depend on the initialization and suffer from the possibility of being trapped in local optima and thus no consistency guarantees can be provided. At each iteration, a large number of candidate structures need to be evaluated, these methods assume that all observed nodes are the leaves of the tree and to reduce the number of candidate structures. Algorithms  have been proposed \citep{Hsu&etal:09COLT} with sample complexity  guarantees for learning HMMs under the condition that the joint distribution of the observed variables generated by distinct hidden states are  distinct.  

The reconstruction of latent trees has   been studied extensively by the {\em phylogenetic} community  where sequences of extant species are available and the unknown phylogenetic  tree is to be inferred from these sequences. See \cite{Durbin} for a thorough overview.   Efficient algorithms with provable performance guarantees are  available \citep{erdos99,daskalakis06}. However, the works in this area mostly assume that only the leaves are observed and each internal node (which is hidden) has the same degree except for the root.  The most popular algorithm for constructing phylogenetic trees is the {\em neighbor-joining (NJ) method} by \cite{Sai87}. 
Like recursive grouping, the input to the algorithm is a set of statistical distances between observed variables.  The algorithm proceeds by recursively pairing two nodes that are the closest neighbors in the true latent tree and introducing a hidden node as the parent of the two nodes.  For more details on NJ, the reader is referred to   \citet[Section~7.3]{Durbin}.

Another popular class of reconstruction methods used in the phylogenetic community is the family of {\em quartet-based  distance methods} \citep{Ban86, erdos99,Jia01}.\footnote{A {\em quartet} is simply an unrooted binary tree on a set of four observed nodes. }  Quartet-based methods first construct a set of quartets for all subsets of four observed nodes. Subsequently, these quartets are then combined  to form a latent tree.  However, when we only have access to the samples at the observed nodes, then it is not straightforward to construct a latent tree from a set of quartets since the   quartets may be not be consistent.\footnote{The term {\em consistent} here is not the same as the estimation-theoretic one. Here, we say that a set of quartets is {\em consistent} if there exists a latent tree such that all quartets agree with the tree.} In fact, it is known that the problem of determining a latent tree that agrees with the maximum number of quartets is NP-hard \citep{Ste92} but many heuristics have been proposed \citep{Far72,Sat77}. Also, when only samples are available, quartet-based methods are usually much less accurate than NJ \citep{StJohn03}  so we only compare our proposed algorithms to NJ. For  further comparisons between (the sample complexity and other aspects of) quartet methods and NJ, the reader is referred to  \cite{Csu00} and \cite{StJohn03}. 

Another distance-based algorithm was proposed in \citet[Section~8.3.3]{pearl88}. This algorithm is very similar in spirit to quartet-based methods  but instead of finding   quartets for {\em all} subsets of four observed nodes, it finds {\em just enough} quartets  to determine the location of each observed node in the tree.  Although the algorithm is consistent, it performs poorly when only the samples of observed nodes are available  \citep[Section~8.3.5]{pearl88}.


The learning of phylogenetic trees is related to the emerging field of {\em network tomography} \citep{Castro04} in which one seeks to learn   characteristics (such as structure) from data which are only available at the end points (e.g., sources and sinks) of the network. However, again observations are only available at the leaf nodes and usually the objective is to estimate the delay distributions corresponding to nodes linked by an edge \citep{Tsang03, Bha09}.   The modeling of the delay distributions is different from the learning of latent tree graphical models discussed in this paper.

\subsection{Paper Organization}
The rest of the paper is organized as follows.  In Section \ref{sec:model}, we introduce the notations and terminologies used in the paper.  In Section~\ref{sec:infoDist}, we introduce the notion of information distances which are used to reconstruct tree models. In the following two sections, we make two assumptions: Firstly, the true distribution is a latent tree and secondly, perfect knowledge of information distance of observed variables is available. We introduce recursive grouping in Section \ref{sec:rg_exact}. This is followed by  our second algorithm  CLGrouping in Section \ref{sec:chowLiu_exact}. In Section~\ref{sec:rg_estimate}, we relax the assumption that the information distances are known and develop sample based algorithms and at the same time provide sample complexity guarantees for recursive grouping and CLGrouping.  We also discuss extensions of our algorithms for the case when the underlying model is not a tree and our goal is to learn an approximation to it using a latent tree model.  We demonstrate the empirical performance of our algorithms in Section \ref{sec:simulations} and conclude the paper in Section \ref{sec:conclusion}.  The Appendix includes proofs for the theorems presented in the paper.


\section{Latent Tree Graphical Models}\label{sec:model}
 
\subsection{Mathematical Notation}\label{subsec:notation} 
Let $G= (W,E)$ be an undirected  graph with vertex (or node) set
$W =\{1,\ldots,M\}$ and edge set $E \subset \binom{W}{2}$. Let
$\nbd(i;G)$ and $\nbd[i;G]$ be the set of neighbors of node $i$ and the {\em closed neighborhood} of $i$ respectively, i.e., $\nbd[i;G]:=\nbd(i;G)\cup\{i\}$.   For a tree  $T=(W,E)$,  the set of leaf nodes (nodes with degree 1), the maximum degree, and the diameter  are denoted by   $\Leaf(T)$, $\Delta(T)$, and  $\diam(T)$ respectively. The {\em path} between two nodes $i$ and $j$ in a tree $T=(W,E)$ is the set of edges connecting $i$ and $j$ and is denoted  as $\Path((i,j); E)$. The {\em depth} of a node in a tree is the shortest distance  (number of hops) to the leaf nodes in $T$. The \emph{parent} of a node $i$ that has depth $s$ is the neighbor of $i$ with depth $s+1$. The set of \emph{child nodes} of a node $i$ with depth $s$ is the set of neighbors of $i$ with depth $s-1$, which we denote as $\calC(i)$.   A set of nodes that share the same parent is called a \emph{sibling} group. A \emph{family} is the union of the siblings and the associated parent.   

A {\em latent tree} is a tree with node set  $W:=V\cup H$,  the union of a set of  observed nodes $V$ (with $m=|V|$), and a set of latent (or hidden) nodes $H$.  The {\em effective depth} $\Depth(T;V)$   (with respect to   $V$) is the maximum distance of a hidden node to its closest observed node, i.e., 
\begin{equation}
\Depth(T;V):=\max_{i\in H} \min_{j\in V} |\Path((i,j); T)|. \label{eqn:depth}
\end{equation}

\subsection{Graphical Models}

An {\em undirected graphical model}~\citep{Lauritzen:book}   is a family of multivariate  probability distributions that factorize according to a graph $G=(W,E)$. More precisely, let $\bX = (X_1,\ldots, X_M)$ be a random vector, where each random variable $X_i$, which takes on values in an alphabet $\calX$,  corresponds to variable at node $i\in V$. The set of edges $E$ encodes the set of conditional independencies in the model. The random vector $\bX$ is said to be {\em Markov} on $G$ if for every   $i$, the random variable $X_i$ is conditionally independent of all other variables  given its neighbors, i.e, if $p$ is the joint distribution\footnote{We abuse the term {\it distribution} to mean a probability mass function in the discrete case (density with respect to the counting measure) and a probability density function (density with respect to the Lebesgue measure) in the continuous case.}  of $\bX$, then 
\begin{equation}
p(x_i|x_{\nbd(i;G)}) = p(x_i|x_{\setminus i}), \label{eqn:local_Markov}
\end{equation}
where $x_{\setminus i}$ denotes the set of all variables\footnote{We will use the terms node, vertex and variable interchangeably in the sequel.  } excluding $x_i$. Eqn.~\eqref{eqn:local_Markov} is known as the {\em local Markov property}. 

In this paper, we consider both  discrete  and Gaussian  graphical models. For discrete models, the alphabet $\calX=\{1,\ldots, K\}$ is  a finite set.  For Gaussian graphical models, $\calX=\bR$ and furthermore, without loss of generality, we assume that the mean is known to be the zero vector and hence, the joint distribution $p(\bx) \propto \exp(-\frac{1}{2} \bx^T \bSigma^{-1} \bx)$ depends only on the covariance matrix $\bSigma$. 

An important and tractable class of graphical models is the set of tree-structured graphical models, i.e., multivariate probability distributions that are Markov on an undirected tree $T=(W,E)$. It is known from junction tree theory \citep{Cow99} that the joint distribution $p$ for such a model factorizes as 
\begin{equation}
p(x_1,\ldots , x_M) = \prod_{i\in W}p(x_i) \prod_{(i,j)\in E} \frac{p(x_i,x_j)}{p(x_i)p(x_j)}. \label{eqn:tree_factor}
\end{equation}
That is, the sets of marginal $\{p(x_i) :i\in W\}$ and pairwise joints on the edges $\{p(x_i, x_j) : (i,j) \in E\}$ fully characterize the joint distribution of a tree-structured graphical model.

A special class of a discrete tree-structured graphical models  is the set of \emph{symmetric discrete distributions}. This class of models is characterized by the fact that the pairs of variables $(X_i,X_j)$ on all the edges $(i,j)\in E$ follow the conditional probability law: 
\begin{equation}
p(x_i|x_j) = \left\{ \begin{array}{ll}
1-(K-1) \theta_{ij}, & \mathrm{if}~x_i=x_j,\\
\theta_{ij}, & \mathrm{otherwise},
\end{array}
\right.
\label{eq:crossover}
\end{equation}
and the marginal distribution of {\em every} variable in the tree is  uniform,  i.e., $p(x_i) = 1/K$ for all $x_i\in \calX$ and for all $i\in V\cup H$. The parameter $\theta_{ij} \in (0, 1/K)  $ in \eqref{eq:crossover}, which does not depend on the states $x_i,x_j\in \calX$,  is known as the {\em crossover probability}. 

Let $\bx^n:=\{\bx^{(1)},\ldots,\bx^{(n)}\}$ be a set of  $n$ i.i.d.\ samples drawn from a graphical model (distribution) $p$, Markov on a latent  tree $T_p=(W ,E_p)$, where $W=V\cup H$.  Each sample $\bx^{(l)}\in \calX^M$ is a length-$M$ vector. In our setup, the learner only has access to samples drawn from the observed node set $V$, and we denote this  set of sub-vectors   containing only the elements in $V$, as $\bx^n_{V} :=\{\bx^{(1)}_{V},\ldots,\bx^{(n)}_{V}\}$, where each observed sample $\bx_V^{(l)}\in \calX^m$ is a length-$m$ vector.  Our algorithms learn latent tree structures using the information distances (defined in Section \ref{sec:infoDist}) between pairs of observed variables, which can be estimated from samples.

\subsection{Minimal Tree Extensions}\label{sec:identifiable}
Our ultimate goal is to recover the graphical model $p$, i.e., the latent tree structure and its parameters,  given  $n$ i.i.d.\ samples of the observed variables $\bx^n_{V}$.  However, in general, there can be multiple latent tree models which result in the same observed statistics, i.e., the same joint distribution   $p_V$ of the observed variables.  We consider the class of  tree models where it is possible to recover the latent tree model uniquely and provide necessary conditions for structure   identifiability, i.e., the identifiability of the edge set $E$. 
 
Firstly, we limit ourselves to the scenario where {\em all} the random variables (both observed and latent) take values on a common alphabet  $\calX$.  Thus, in the Gaussian case, each hidden and observed variable is a univariate Gaussian. In the discrete case, each variable  takes on values in the same finite alphabet $\calX$. Note that the model may not be identifiable if some of the hidden variables are allowed  to have arbitrary  alphabets. As an example, consider a discrete latent tree model with binary observed variables ($K=2$). A latent tree with the simplest structure (fewest number of nodes) is a tree in which all $m$  observed binary variables are connected to one hidden variable.  If we allow the hidden variable to take on $2^m$ states, then the tree can describe all possible statistics among the $m$ observed variables, i.e., the joint distribution $p_V$ can be arbitrary.\footnote{This follows from a elementary parameter counting argument. }    

A probability distribution $p_V (\bx_V)$ is said to be \emph{tree-decomposable} if it is the marginal (of variables in $V$) of a tree-structured graphical model $p (\bx_V, \bx_H)$.  In this case,  $p$ (over variables in $W$) is said to be a \emph{tree extension} of $p_V$ \citep{pearl88}.  A distribution $p$  is said to have a \emph{redundant} hidden node $h\in H$ if we can remove $h$ and the marginal on the set of visible nodes $V$ remains as $p_V$.  The following conditions ensure that a latent tree does not include a redundant hidden node \citep{pearl88}:

\begin{itemize}
\item[(C1)] 
 Each hidden variable has at least three neighbors (which can be either hidden or observed).  Note that this ensures that all leaf nodes are observed (although not all observed nodes need to be leaves). 
\item[(C2)]  Any two variables connected by an edge in the tree model  are neither  perfectly dependent nor    independent. 
\end{itemize}
  
\begin{figure}[t]
\begin{center}
\includegraphics[width=0.4\linewidth]{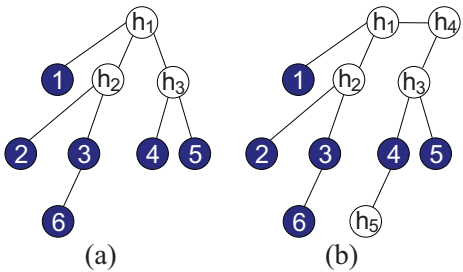} 
\end{center}
\caption{Examples of minimal latent trees.  Shaded nodes are observed and unshaded nodes are hidden.  (a) An identifiable tree. (b) A non-identifiable tree because $h_4$ and $h_5$ have degrees less than $3$.}
\label{fig:identifiable}
\end{figure}

Figure \ref{fig:identifiable}(a) shows an example of a tree satisfying (C1).  If (C2), which is a condition on parameters, is also satisfied, then the tree in Figure \ref{fig:identifiable}(a) is identifiable.  The tree shown in Figure \ref{fig:identifiable}(b) does not satisfy (C1) because $h_4$ and $h_5$ have degrees less than $3$.  In fact, if we marginalize out the hidden variables $h_4$ and $h_5$, then the resulting model has the same tree structure as in Figure \ref{fig:identifiable}(a).

We assume throughout the paper that (C2) is satisfied for all probability distributions. Let $\calT_{\geq 3}$ be the set of (latent) trees satisfying (C1). We refer to $\calT_{\ge 3}$ as the set of {\em minimal (or identifiable) latent trees}. Minimal latent trees do not contain redundant hidden nodes.  The   distribution $p$ (over $W$ and Markov on some tree in $\calT_{\ge 3}$) is said to be a \emph{minimal tree extension} of $p_V$. As illustrated in Figure~\ref{fig:identifiable}, using marginalization operations, any non-minimal latent tree distribution can be reduced to a minimal latent tree model. 

{\PROP {\bf (Minimal Tree Extensions) \citep[Section 8.3]{pearl88}} \label{prop:identifiability}  
\begin{enumerate}
\item[(i)] For every tree-decomposable distribution $p_V$, there exists a minimal tree extension $p$ Markov on a tree $T \in \calT_{\geq 3}$, which is unique up to the renaming of the variables or their values. 
\item[(ii)] For Gaussian and binary distributions, if $p_V$ is known exactly, then the minimal tree extension $p$ can be recovered. 
\item[(iii)] The structure of $T$ is uniquely determined by the pairwise   distributions of observed variables $p(x_i,x_j)$ for all $i,j\in V$. \end{enumerate}
}


\subsection{Consistency}
We now define the notion of consistency. In Section~\ref{sec:rg_estimate}, we show that our latent tree learning algorithms are consistent. 

{\DEFI {\bf (Consistency)} \label{def:consistent} A {\em latent tree reconstruction algorithm} $\calA$ is a map from the observed samples  $\bx_V^n$ to an estimated  tree $\hT^n$ and an estimated tree-structured graphical model $\hp^n$. We say that a latent tree reconstruction algorithm $\calA$ is {\em structurally consistent} if  there exists a graph homomorphism\footnote{A graph homomorphism is a mapping between graphs that respects their structure. More precisely, a {\em graph homomorphism} $h$ from a graph $G=(W,E)$ to a graph $G'=(V',E')$, written $h:G\to G'$ is a mapping $h:V\to V'$ such that $(i,j)\in E$ implies that $(h(i),h(j))\in E'$.} $h$ such that 
\begin{equation}
\lim_{n\to\infty} \Pr ( h(\hT^n)\neq T_p )=0.
\label{eqn:structural_consistency}
\end{equation} 
Furthermore, we say that  $\calA$ is {\em risk consistent} if to every $\varepsilon>0$,
\begin{equation}
\lim_{n\to\infty} \Pr \left( D(p \, ||\, \hp^n) >\varepsilon  \right)=0,\label{eqn:consistency}
\end{equation}
where $D(p\, ||\, \hp^n)$ is the KL-divergence \citep{Cov06} between the true distribution $p$ and the estimated distribution $\hp^n$. 
}

\vspace{0.05in} 


In the following sections, we   design structurally  and risk consistent algorithms   for (minimal) Gaussian and symmetric discrete latent tree models, defined in~\eqref{eq:crossover}. Our algorithms use pairwise distributions  between the observed nodes. However, for general discrete models, pairwise distributions between observed nodes   are, in general, not sufficient to recover the parameters \citep{Cha91}. Therefore, we only prove structural consistency, as defined in~\eqref{eqn:structural_consistency}, for general discrete latent tree models.   For such distributions, we consider a two-step procedure for structure and parameter estimation: Firstly, we estimate the structure of the latent tree using the algorithms suggested in this paper. Subsequently, we use the Expectation Maximization (EM) algorithm~\citep{dempster77} to infer the parameters. Note that, as mentioned, risk consistency will not be guaranteed in this case.

\section{Information Distances}
\label{sec:infoDist}
 
The proposed algorithms in this paper receive as inputs the set of so-called  (exact or estimated) {\em information distances}, which are functions of the pairwise distributions. These quantities  are defined in Section~\ref{sec:infodistdef} for the two classes of tree-structured graphical models discussed in this paper, namely the Gaussian and discrete graphical models. We also show that the information distances have a particularly simple form for symmetric discrete distributions.  In Section~\ref{sec:testing_node}, we use the information distances to infer the relationships between the observed variables such as $j$ is a child of $i$ or $i$ and $j$ are siblings. 
 
\subsection{Definitions of  Information Distances} \label{sec:infodistdef}
We define \emph{information distances} for Gaussian and discrete distributions and show that these distances are additive for tree-structured graphical models. Recall that for two random variables $X_i$ and $X_j$, the  {\em correlation coefficient} is defined as 
\begin{equation}
\rho_{ij} := \frac{\mathrm{Cov} (X_i,X_j)}{\sqrt{ \mathrm{Var}(X_i) \mathrm{Var}(X_j)  }}. \label{eqn:corr_coeff}
\end{equation}
\noindent For Gaussian graphical models, the information distance associated with the pair of variables $X_i$ and $X_j$ is defined as:
\begin{equation}
d_{ij}:=-\log |\rho_{ij}|.
\label{eq:dist_gaussian}
\end{equation}
Intuitively, if the information distance $d_{ij}$ is large, then $X_i$ and $X_j$ are weakly correlated and vice-versa. 

For discrete graphical models, let $\mathbf{J}^{ij}$ denote the joint probability matrix between $X_i$ and $X_j$ (i.e., ${J}^{ij}_{ab} = p(x_i = a, x_j = b),a,b\in\calX$). Also let $\mathbf{M}^i$ be the diagonal marginal probability matrix of $X_i$ (i.e., $M_{aa}^{i} = p(x_i = a)$).  For discrete graphical models, the information distance associated with the pair of variables $X_i$ and $X_j$ is defined as \citep{lake94}:
\begin{equation}
d_{ij} := -\log \frac{|\det \mathbf{J}^{ij}|}{\sqrt{\det \mathbf{M}^i \det \mathbf{M}^j}}. 
\label{eq:dist_discrete}
\end{equation}
Note that for binary variables, i.e., $K=2$, the value of $d_{ij} $ in~\eqref{eq:dist_discrete} reduces to the expression in~\eqref{eq:dist_gaussian}, i.e., the information distance is   a function of the correlation coefficient, defined in~\eqref{eqn:corr_coeff}, just as in the Gaussian case.

For  symmetric discrete distributions defined in \eqref{eq:crossover}, the information distance defined for discrete graphical models in~\eqref{eq:dist_discrete} reduces to 
\begin{equation}
d_{ij} := -(K-1)\log (1 - K \theta_{ij}).
\label{eq:dist_symdiscrete}
\end{equation}
Note that there is one-to-one correspondence between the information distances $d_{ij}$ and model parameters for  Gaussian distributions (parametrized by the correlation coefficient $\rho_{ij}$) in~\eqref{eq:dist_gaussian} and the  symmetric discrete distributions (parametrized by  the crossover probability $\theta_{ij}$) in~\eqref{eq:dist_symdiscrete}. This is, however, not true for general discrete distributions. 

Equipped with these definitions of information distances, assumption (C2) in Section~\ref{sec:identifiable} can be rewritten as the following: There exists constants $0<l,u <\infty$, such that 
\begin{equation}
\mbox{(C2)}\qquad l\leq d_{ij}\le u, \qquad\forall\, (i,j)\in E_p. \label{eqn:infodistbounds}
\end{equation}

{\PROP\label{fact:mst} {\bf (Additivity of Information Distances)} The information distances $d_{ij}$ defined in (\ref{eq:dist_gaussian}), (\ref{eq:dist_discrete}), and (\ref{eq:dist_symdiscrete}) are {\em additive tree metrics}  \citep{erdos99}.  In other words, if the joint probability distribution $p(\bx)$ is a tree-structured graphical model Markov on the tree $T_p=(W,E_p)$, then the information distances are additive on $T_p$:\begin{equation}
d_{kl} = \sum_{(i,j) \in \Path((k,l); E_p)} \, d_{ij},\quad \forall k,l\in W.\label{eqn:markov}
\end{equation}   }
\vspace{0.05in} 

The   property in~\eqref{eqn:markov}  means that  if each pair of vertices $i,j \in W$ is assigned the weight $d_{ij}$, then $T_p$ is a minimum spanning tree on $W$, denoted as $\MST(W;\bD)$, where $\bD$ is the information distance matrix with elements $d_{ij}$ for all $ i,j\in V$.

It is straightforward to show that the information distances are additive for the Gaussian and symmetric discrete cases using the local Markov property of graphical models.  For general discrete distributions with information distance as in \eqref{eq:dist_discrete}, see \cite{lake94} for the proof. In the rest of the paper, we map the parameters of Gaussian and discrete distributions to an information distance matrix $\bD=[d_{ij}]$ to unify the analyses for both cases.  

\subsection{Testing Inter-Node Relationships} \label{sec:testing_node}
In this section, we use Proposition~\ref{fact:mst} to ascertain child-parent and sibling (cf.\ Section~\ref{subsec:notation}) relationships between the variables in a latent tree-structured graphical model. To do so,  for any three variables $i,j,k\in V$, we define  $\Phi_{ijk} := d_{ik}-d_{jk}$ to be the difference between the information distances $d_{ik}$ and $d_{jk}$.  The following lemma suggests a simple procedure to identify the set of relationships between the nodes. 


{\LEMM  {\bf (Sibling Grouping)}  \label{th:gauss_family}  
For distances $d_{ij}$  for all $i,j\in V$ on a tree $T \in \calT_{\ge 3}$,  the following  two properties on $\Phi_{ijk} = d_{ik} - d_{jk}$ hold:
\begin{itemize}
\item[(i)] $\Phi_{ijk} = d_{ij}$ for all $k \in V\setminus \{i,j\}$ if and only if $i$ is a leaf node and $j$ is its parent. 
\item[(ii)]  $-d_{ij}  <  \Phi_{ijk}  =  \Phi_{ijk'}  <  d_{ij}$ for all $k,k'  \in V\setminus \{i,j\}$ if and only if both $i$ and $j$ are leaf nodes and they have the same parent, i.e., they belong to the same sibling group.
\end{itemize}
}
\vspace{0.05in} 

\begin{figure}[t]
\centerline{\includegraphics[width=\linewidth]{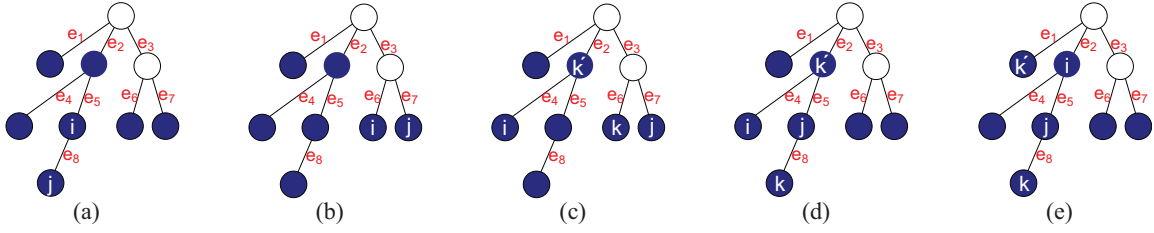}}
\caption{Examples for each case in $\TestNodeRelationships$. For each edge, $e_i$ represents the information distance associated with the edge.  (a) Case 1: $\Phi_{ijk}=-e_8 = -d_{ij} $ for all $k\in V\setminus\{i,j\}$. (b) Case 2: $\Phi_{ijk} = e_6 - e_7 \neq d_{ij} = e_6 + e_7$ for all $k\in V\setminus\{i,j\}$ (c) Case 3a: $\Phi_{ijk} = e_4+e_2 + e_3 - e_7 \neq \Phi_{ijk'} = e_4 - e_2 - e_3 - e_7$. (d) Case 3b: $\Phi_{ijk} = e_4+e_5 \neq \Phi_{ijk'} = e_4 - e_5$. (e) Case 3c: $\Phi_{ijk} = e_5 \neq \Phi_{ijk'} = -e_5 $.}
\label{fig:TestNodeRelationships}
\end{figure}

The proof of the lemma uses Proposition \ref{fact:mst} and is provided in Appendix~\ref{app:th:gauss_family}. Given Lemma~\ref{th:gauss_family}, we can first determine all the values of $\Phi_{ijk}$ for triples $i,j,k\in V$. Now we can determine the relationship between nodes $i$ and $j$ as follows: Fix the pair of nodes $i,j\in V$ and consider all the other nodes $k\in V\setminus \{i,j\}$. Then,  there are three possibilities for the set  $\{\Phi_{ijk}: k\in V\setminus\{i,j\}\}$:
\begin{enumerate}
\item $\Phi_{ijk}=d_{ij}$  for all $k\in V\setminus\{i,j\}$. Then, $i$ is a leaf node and $j$ is a parent of $i$.  Similarly, if $\Phi_{ijk}=-d_{ij}$  for all $k\in V\setminus\{i,j\}$, $j$ is a leaf node and $i$ is a parent of $j$. 
\item $\Phi_{ijk}$ is constant  for all $k\in V\setminus\{i,j\}$ but {\em not} equal to either $d_{ij}$ or $-d_{ij}$. Then $i$ and $j$ are leaf nodes and they are  siblings. 
\item $\Phi_{ijk}$ is not equal for all $k\in V\setminus\{i,j\}$. Then, there are three possibilities: Either
\begin{enumerate}
\item Nodes $i$ and $j$ are not siblings nor have a parent-child relationship or,
\item Nodes $i$ and $j$ are siblings but at least one of them is not a leaf or,
\item Nodes $i$ and $j$ have a parent-child relationship but the child is not a leaf. 
\end{enumerate}
\end{enumerate}
Thus, we have a simple test to determine the relationship between $i$ and $j$ and to ascertain whether $i$ and $j$ are leaf nodes. We call the above test $\TestNodeRelationships$. See Figure~\ref{fig:TestNodeRelationships} for examples.  By running this test for all $i$ and $j$, we can determine all the relationships among all pairs of observed variables. 

In the following section, we describe a recursive algorithm that is based on the above  $\TestNodeRelationships$ procedure  to reconstruct the entire latent tree model assuming that the true distance matrix $\bD=[d_{ij}]$ are known. In Section~\ref{sec:chowLiu_exact}, we provide improved algorithms for the learning of latent trees again assuming that $\bD$ is known. Subsequently, in Section~\ref{sec:rg_estimate}, we develop algorithms for the  {\em consistent} reconstruction of latent trees when information  distances   are unknown  and we have to estimate them from the samples $\bx_{V}^n$. In addition, in Section~\ref{subsec:regCLG}  we discuss how to extend these algorithms for the case when    $p_V$ is not  necessarily tree-decomposable, i.e., the original graphical model is not assumed to be a latent tree.


%
%


\section{Recursive Grouping  Algorithm Given Information Distances}
\label{sec:rg_exact}

This section is devoted to the development  of the first algorithm for reconstructing latent tree models, recursive grouping (RG). At a high level, RG is a recursive procedure in which at each step,  $\TestNodeRelationships$   is used  to identify  nodes that belong to the same family. Subsequently,  RG introduces a parent node if a family  of nodes (i.e., a sibling group) does not contain an observed parent. This newly introduced parent node corresponds   to a hidden node in the original unknown latent tree. Once such a parent (i.e.,  hidden) node $h$ is introduced, the information distances from $h$ to all other observed nodes can be computed. 



The inputs to RG are the vertex set $V$ and the matrix of information distances $\bD$ corresponding to a latent tree.  The algorithm proceeds by recursively grouping nodes and adding hidden variables. In each iteration, the algorithm acts on a so-called active set of nodes $Y$, and in the process constructs a new active set $Y_{\new}$ for the next iteration.\footnote{Note that the current active set is also used (in Step \ref{item:recompute}) after the new active set has been defined. For clarity, we also introduce the quantity $Y_{\old}$ in Steps \ref{item:updateY} and \ref{item:recompute}.}  The steps are as follows:
\begin{enumerate}
\item Initialize by setting $Y:=V$ to be the set of observed variables. 
\item Compute $\Phi_{ijk}=d_{ik} -d_{jk} $ for all $i,j,k\in Y$.
\item  \label{item:partitions} Using the $\TestNodeRelationships$ procedure, define $\{\Pi_l\}_{l=1}^L$ to be the coarsest partition\footnote{Recall that a {\em partition} $P $ of a set $Y$ is a collection of nonempty subsets $\{\Pi_l\subset Y\}_{l=1}^L$ such that $\cup_{l=1}^L \Pi_l=Y $ and $\Pi_l\cap \Pi_{l'}=\emptyset$ for all $l\ne l'$. A partition $P$ is said to be  {\em coarser than}   another partition $P'$ if every element of $P'$ is a subset of some element of $P$.  } of $Y$ such that for every subset $\Pi_l$ (with $|\Pi_l|\ge 2$), any two nodes in  $\Pi_l$ are either siblings which are leaf nodes or they have a parent-child relationship in which the child is a leaf. Note that for some $l$, $\Pi_l$ may consist of a single node.  Begin to construct the new active set by setting $Y_{\new} \leftarrow \bigcup_{l:|\Pi_l| = 1}  \Pi_l$.
\item \label{item:notcontain} For each $l=1,\ldots, L$ with $|\Pi_l| \ge 2$, if  $\Pi_l$ contains a parent node $u$, update $Y_{\new} \leftarrow Y_{\new} \cup \{u\}$.  Otherwise, introduce a new hidden node $h$, connect $h$ (as a parent) to every node in $\Pi_l$, and set $Y_{\new} \leftarrow Y_{\new} \cup \{h\}$.
\item  \label{item:updateY} Update the active set: $Y_{\old} \leftarrow Y$ and $Y \leftarrow Y_{\new}$.

\item \label{item:recompute} 
For each new hidden node $h \in Y$, compute the information distances $d_{hl}$ for all $l \in Y$ using~\eqref{eq:dist_childparent_exact} and~\eqref{eq:dist_hiddenother_exact} described below. 
\item If $|Y|\ge 3$, return to step 2. Otherwise, if $|Y|=2$, connect the two remaining nodes in $Y$ with an edge then stop. If instead $|Y|=1$, do nothing and stop. 
\end{enumerate}

We now describe how to compute the information distances in Step ~\ref{item:recompute}  for each new hidden node $h \in Y$ and all other active nodes $l\in Y$. Let $i,j \in \calC(h)$ be two children of $h$, and let $k \in Y_{\old} \setminus \{ i,j \}$ be any other node in the previous active set. From Proposition~\ref{fact:mst}, we have that $d_{ih}-d_{jh}=d_{ik} -d_{jk}  = \Phi_{ijk}$ and $d_{ih}+d_{jh}=d_{ij} $, from which we can recover the information distances between a previously active node $i \in Y_{\old}$  and its new hidden parent $h \in Y$ as follows:
\begin{equation}
d_{ih} = \frac{1}{2} \left(d_{ij}  + \Phi_{ijk} \right).
\label{eq:dist_childparent_exact}
\end{equation}
For any other active node $l \in Y$, we can compute $d_{hl} $ using a child node $i \in \calC(h)$ as follows:
\begin{equation}
d_{hl} = \left\{
\begin{array}{ll}
d_{il} - d_{ih}, & \mathrm{if}~l \in Y_{\old},\\
d_{ik} - d_{ih} - d_{lk}, &\mathrm{otherwise,~where~} k \in \calC(l).
\end{array}
\right.
\label{eq:dist_hiddenother_exact}
\end{equation}
Using equations \eqref{eq:dist_childparent_exact} and \eqref{eq:dist_hiddenother_exact}, we can infer all the information distances $d_{hl}$ between a newly introduced hidden node $h$ to all other active nodes $l \in Y$. Consequently, we have all the distances $d_{ij} $ between all pairs of nodes in the active set $Y$. It can be shown that this algorithm recovers all minimal latent trees. The proof of the following theorem is provided in Appendix~\ref{app:th:correct_rg}.



{\THEO {\bf (Correctness and Computational Complexity of RG)} \label{th:correct_rg} If $T_p\in\calT_{\geq 3}$ and the matrix of information distances $\bD$ (between nodes in $V$) is available, then RG   outputs the true latent tree $T_p$ correctly in time  $O(\diam(T_p) m^3)$.}
\vspace{0.05in}

\begin{figure}[t]
\begin{center}
\includegraphics[width=0.8\linewidth]{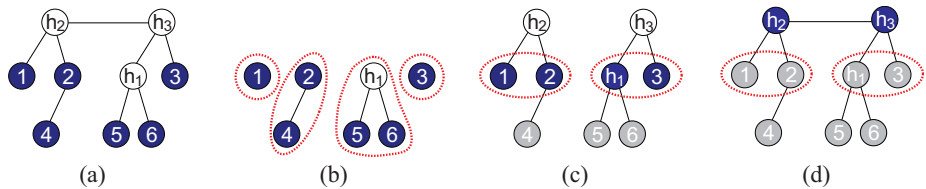} 
\end{center}
\caption{An illustrative example of RG. Solid nodes indicate the active set $Y$ for each iteration.  (a) Original latent tree. (b) Output after the first iteration of RG.  Red dotted lines indicate the subsets  $\Pi_l$ in the partition of $Y$.  (c) Output after the second iteration of RG.  Note that $h_3$, which was introduced in the first iteration, is an active node for the second iteration.  Nodes 4,5, and 6 do not belong to the current active set and are represented in grey.  (d) Output after the third iteration of RG, which is same as the original latent tree.}
\label{fig:rg_example}
\end{figure}


We now use a concrete example to illustrate the steps involved in RG. In Figure~\ref{fig:rg_example}(a), the original unknown latent tree is shown. In this tree, nodes $1,\ldots, 6$ are the observed nodes and $h_1,h_2,h_3$ are the hidden nodes. We start by considering the set of observed nodes as active nodes $Y:=V=\{1,\ldots, 6\}$. Once $\Phi_{ijk}$ are computed from the given distances $d_{ij} $, $\TestNodeRelationships$ is used to determine that   $Y$ is  partitioned into four subsets: $\Pi_1 = \{1\}, \Pi_2 = \{2,4\} , \Pi_3=\{5,6\}, \Pi_4=\{3\}$. The subsets $\Pi_1$ and $\Pi_4$ contain only one node. The subset $\Pi_3$ contains two siblings that are leaf nodes. The subset $\Pi_2$ contains a parent node 2 and a child node 4, which is a leaf node. Since $\Pi_3$ does not contain a parent, we introduce a new hidden node $h_1$ and connect $h_1$ to 5 and 6 as shown in Figure~\ref{fig:rg_example}(b). The information distances $d_{5h_1}$ and $d_{6h_1}$ can be computed using~\eqref{eq:dist_childparent_exact}, e.g., $d_{5h_1} = \frac{1}{2}(d_{56} + \Phi_{561})$. The new active set is the union of all nodes in the single-node subsets, a parent node, and a new hidden node $Y_{\new}=\{1,2,3,h_1\}$. Distances among the pairs of nodes in $Y_{\new}$ can be computed using~\eqref{eq:dist_hiddenother_exact} (e.g., $d_{1h_1} = d_{15}-d_{5h_1} $). In the second iteration, we again use $\TestNodeRelationships$ to ascertain that $Y$ can be partitioned into $\Pi_1=\{1,2\}$ and $\Pi_2=\{h_1,3\}$. These two subsets do not have parents so $h_2$ and $h_3$ are added to $\Pi_1$ and $\Pi_2$ respectively. Parent nodes $h_2$ and $h_3$ are connected to their children in $\Pi_1$ and $\Pi_2$ as shown in Figure~\ref{fig:rg_example}(c). Finally, we are left with the active set as $Y=\{h_2,h_3\}$ and the algorithm terminates after $h_2$ and $h_3$ are connected by an edge. The hitherto unknown latent tree is fully reconstructed as shown in Figure~\ref{fig:rg_example}(d).

A potential drawback of RG is that it involves multiple {\em local} operations, which may   result in a high computational complexity. Indeed, from Theorem~\ref{th:correct_rg}, the worst-case complexity  is $O(m^4)$ which occurs when $T_p$, the true latent tree, is a hidden Markov model (HMM). This may be computationally prohibitive if $m$ is large. In Section~\ref{sec:chowLiu_exact} we design an algorithm which uses a  {\em global} pre-processing step to  reduce the overall   complexity substantially, especially for trees with large diameters (of which HMMs are extreme examples).

\section{CLGrouping Algorithm Given Information Distances}
\label{sec:chowLiu_exact} 
In this section, we present CLGrouping, an algorithm for reconstructing latent trees more efficiently than RG. As in Section~\ref{sec:rg_exact}, in this section, we assume that $\bD$ is known; the extension to unknown $\bD$ is discussed in Section~\ref{subsec:clgroup_estimate}.   CLGrouping   is a two-step procedure, the first of which is  a global pre-processing step that involves the construction of a  so-called {\em Chow-Liu tree} \citep{chow68} over the set of observed nodes $V$. This   step identifies  nodes that do not belong to the same sibling group.  In the second step, we complete the recovery of the latent tree by applying a distance-based latent tree reconstruction algorithm (such as RG or NJ) repeatedly on smaller subsets of nodes.  We review the Chow-Liu algorithm in Section~\ref{sec:cl}, relate the Chow-Liu tree to the true latent tree in Section~\ref{sec:cl_latent}, derive a simple transformation of the Chow-Liu tree to obtain the latent tree in Section~\ref{sec:blind} and propose CLGrouping in Section~\ref{sec:clgrouping}.  For simplicity, we focus on the Gaussian distributions and the symmetric discrete distributions first, and discuss the extension to general discrete models in Section~\ref{sec:general_discrete}.



\subsection{A Review of the Chow-Liu Algorithm} \label{sec:cl}
In this section, we review the Chow-Liu tree reconstruction procedure. To do so, define $\calT(V)$ to be the set of trees with vertex set $V$ and  $\calP(\calT(V))$ to  be the set of tree-structured graphical models whose graph has vertex set $V$, i.e., every $q \in \calP(\calT(V))$ factorizes as in~\eqref{eqn:tree_factor}. 

Given an arbitrary multivariate distribution $p_V(\bx_V)$, \cite{chow68} considered the following {\em KL-divergence minimization} problem:
\begin{equation}
p_{\CL} :=\argmin_{q\in \calP(\calT(V))} \,\, D(p_V\,||\, q). \label{eqn:q_tree}
\end{equation}
That is, among all the tree-structured graphical models with  vertex set $V$, the distribution $p_{\CL}$ is the closest one to $p_V$ in terms of the KL-divergence. By using the factorization property in \eqref{eqn:tree_factor}, we can easily verify that $p_{\CL}$  is Markov on the {\em Chow-Liu tree} $T_{\CL}=(V,E_{\CL})$ which   is given by the optimization problem:\footnote{In \eqref{eqn:cl_mwst} and the rest of the paper, we adopt the following simplifying notation; If $T=(V,E)$ and if $(i,j)\in E$, we will also say that $(i,j)\in T$. }
\begin{equation}
T_{\CL} = \argmax_{T \in \calT(V) }  \,\,\sum_{(i,j)\in T} I(X_i\,;\, X_j). \label{eqn:cl_mwst}
\end{equation}
In~\eqref{eqn:cl_mwst}, $I(X_i \, ;\, X_j)=D(p(x_i,x_j) \, ||\,  p(x_i)\,  p(x_j))$ is the   {\em mutual information} \citep{Cov06} between random variables $X_i$ and $X_j$. The optimization in~\eqref{eqn:cl_mwst} is a max-weight spanning tree  problem \citep{Cor03} which can be solved efficiently in time $O(m^2\log m)$ using either Kruskal's algorithm \citep{Kruskal}  or Prim's algorithm  \citep{Prim}. The edge weights for the max-weight spanning tree  are precisely the mutual information quantities between random variables. Note that the parameters of $\hp$ in~\eqref{eqn:q_tree} are found by setting the pairwise distributions   $p_{\CL}(x_i, x_j)$ on the edges to  $p_V(x_i, x_j)$, i.e., $p_{\CL}(x_i, x_j) = p_V(x_i, x_j)$ for all $(i,j)\in E_{\CL}$. We now relate the Chow-Liu tree on the observed nodes and the information distance matrix $\bD $.

{\LEMM[Correspondence between $T_{\CL}$ and $\MST$]\label{lemma:dist} 
If $p_V$ is a  Gaussian distribution or a symmetric discrete distribution, then the Chow-Liu tree  in~\eqref{eqn:cl_mwst} reduces to the minimum spanning tree (MST)  where the edge weights are  the information distances $d_{ij} $,  i.e., 
\begin{equation}
T_{\CL} =\MST( V;\bD) := \argmin_{ T\in \calT(V)} \,\, \sum_{(i,j) \in T}\,\, d_{ij} . \label{eqn:TMST}
\end{equation}  }
\vspace{0.05in} 

Lemma~\ref{lemma:dist}, whose proof is omitted, follows because for Gaussian and symmetric discrete models,  the mutual information\footnote{Note that, unlike information distances $d_{ij}$, the mutual information quantities $I(X_i\,;\, X_j)$ do not form an additive metric on $T_p$.} $I(X_i\, ; \, X_j)$  is a monotonically decreasing function of the  information distance $d_{ij} $.\footnote{For example, in the case of Gaussians, $I(X_i\, ; \, X_j) = -\frac{1}{2}\log (1-\rho_{ij}^2)$  \citep{Cov06}. }  For other graphical models (e.g., non-symmetric discrete distributions), this relationship is not necessarily true. See Section~\ref{sec:general_discrete} for a discussion. Note that when all nodes are observed (i.e., $W=V$),   Lemma~\ref{lemma:dist} reduces to Proposition~\ref{fact:mst}. 
 
\subsection{Relationship between the Latent Tree and the Chow-Liu Tree (MST)}
\label{sec:cl_latent}
In this section, we relate $\MST( V;\bD)$ in \eqref{eqn:TMST} to the original latent tree $T_p$.   To relate the two trees, $\MST(V;\bD)$ and $T_p$, we first introduce the notion of a surrogate node. 

{\DEFI ({\bf Surrogate Node}) \label{def:surrogate} 
Given the latent tree $T_p=(W, E_p)$ and any node $i\in W$, the   \emph{surrogate node}  of $i$ with respect to $V$ is defined as   
\begin{equation} \label{eqn:surr}
\Sg(i;T_p,V) := \argmin_{j\in V}   d_{ij} . 
\end{equation} }
\vspace{0.05in} 

Intuitively, the surrogate node of a hidden node  $h\in H$ is an observed node $j\in  V$ that is most strongly correlated to $h$. In other words, the information distance between $h$ and $j$ is the smallest. Note that if $i \in  V$, then $\Sg(i;T_p,V) =  i$ since $d_{ii} =  0$.  The map $\Sg(i;T_p,V)$ is a many-to-one function, i.e., several nodes may have the same surrogate node, and its inverse is the {\em inverse surrogate set of $i$} denoted as 
\begin{equation}
\Sg^{-1}(i;T_p,V) :=  \{h\in W :   \Sg(h;T_p,V)=i\}.
\end{equation}
When the tree $T_p$ and the observed vertex set $V$ are understood from context, the surrogate node of $h$  and the inverse surrogate set of $i$ are  abbreviated as $\Sg(h)$ and $\Sg^{-1}(i)$ respectively. We now relate the original latent tree $T_p=(W,E_p)$ to the Chow-Liu tree (also termed the MST) $\MST(V;\bD)$   formed using the distance matrix $\bD$.  

{\LEMM\label{lemma:surrogate} ({\bf Properties of the MST}) 
The    MST  in \eqref{eqn:TMST} and surrogate nodes satisfy the following properties:
\begin{enumerate}
\item[(i)] The surrogate nodes of any two neighboring nodes in $E_p$ are   neighbors in the MST, i.e., for all $i,j\in W$ with $\Sg(i)\neq \Sg(j)$, \begin{equation} (i,j)\in E_p \Rightarrow (\Sg(i), \Sg(j))\in \MST( V;\bD). \label{eqn:surrogate:item1}\end{equation} 
 \item[(ii)] If $j\in V$ and $h\in\Sg^{-1}(j)$, then every node along the path connecting $j$ and $h$ belongs to the inverse surrogate set $\Sg^{-1}(j)$.  
\item[(iii)] The maximum degree of the   MST satisfies 
\begin{equation}
\Delta(\MST(V;\bD))\le \Delta(T_p)^{1+ \frac{u}{l} \, \Depth(T_p;V)}, \label{eqn:surrogate:item2} 
\end{equation} where $\Depth(T_p;V)$ is the effective depth defined in \eqref{eqn:depth} and $l,u$ are the bounds on the information distances on edges in $T_p$ defined in  \eqref{eqn:infodistbounds}.
\end{enumerate}
}
 
\vspace{0.05in} 

The proof of this result can be found in Appendix~\ref{app:lemma:surrogate}. As a result of Lemma~\ref{lemma:surrogate}, the properties of  $\MST(V;\bD)$ can be expressed in terms of the original latent tree $T_p$. For example, in Figure~\ref{fig:CL_example}(a), a latent tree is shown with its corresponding surrogacy relationships, and Figure~\ref{fig:CL_example}(b) shows the corresponding MST over the observed nodes. 

The properties in Lemma~\ref{lemma:surrogate}(i-ii) can also be regarded as \emph{edge-contraction operations} \citep{Rob81} in the original latent tree to obtain the MST. More precisely, an edge-contraction operation on an edge $(j,h) \in V\times H$ in the latent tree $T_p$ is defined as the ``shrinking'' of $(j,h)$ to a single node whose label is the observed node $j$. Thus, the edge $(j,h)$ is ``contracted'' to a single node $j$. By using Lemma~\ref{lemma:surrogate}(i-ii), we observe that the Chow-Liu tree $\MST(V;\bD)$ is formed by applying edge-contraction operations sequentially to each $(j,h)$ pair for all $h\in \Sg^{-1}(j)\cap H$ until all pairs have been contracted to a single node $j$. For example, the MST in Figure~\ref{fig:CL_example}(b) is obtained by contracting edges $(3,h_3)$, $(5,h_2)$, and then $(5,h_1)$ in the latent tree in Figure~\ref{fig:CL_example}(a).


The properties in Lemma~\ref{lemma:surrogate} can be used to design efficient algorithms based on transforming the MST to obtain the latent tree $T_p$. Note that the maximum degree of the MST, $\Delta(\MST(V;\bD))$, is bounded by the maximum degree  in the original latent tree. The quantity $\Delta(\MST(V;\bD))$  determines the computational complexity of one of our proposed algorithms (CLGrouping) and it is small if the depth of the latent tree $\delta(T_p;V)$ is small and the information distances $d_{ij} $  satisfy tight bounds (i.e., $u/l$ is close to unity). The latter condition holds for (almost) homogeneous models in which all the information distances $d_{ij} $ on the edges are almost equal.


 
\subsection{Chow-Liu Blind Algorithm for a Subclass of Latent Trees}\label{sec:blind}

In this section, we present a simple and intuitive  transformation of the Chow-Liu tree that produces the original latent tree.  However, this algorithm, called Chow-Liu Blind (or CLBlind), is applicable only to a  subset  of latent trees called {\em blind latent tree-structured graphical models} $\calP(\calT_{\blind})$. Equipped with the intuition    from CLBlind,  we generalize  it in Section~\ref{sec:clgrouping} to design the CLGrouping algorithm that produces the correct latent tree structure from the MST  {\em for all}  minimal latent tree models. If $p\in \calP(\calT_{\blind})$, then its structure $T_p=(W,E_p)$ and the distance matrix $\bD$ satisfy the following properties: 
\begin{enumerate}
\item[(i)] The true latent tree $T_p\in\calT_{\ge 3}$ and all the internal nodes\footnote{Recall that an internal node is one whose degree is greater than or equal to 2, i.e., a non-leaf.} are hidden,  i.e., $V = \Leaf(T_p)$. 
\item[(ii)] The surrogate node of (i.e., the observed node with the strongest correlation with) each hidden node is one of its children, i.e., $\Sg(h) \in \calC(h)$ for all $h\in H$.   
\end{enumerate}
\begin{figure}[t]
\includegraphics[width=\linewidth]{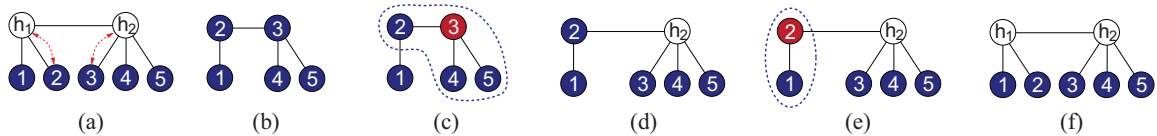} 
\caption{An illustration of CLBlind. The shaded nodes are the observed nodes and the rest are hidden nodes. The dotted lines denote surrogate mappings for the hidden nodes. (a) Original latent tree, which belongs to the class of blind latent graphical models, (b) Chow-Liu tree over the observed nodes, (c) Node 3 is the input to the blind transformation, (d)  Output after the blind transformation, (e) Node 2 is the input to the blind transformation, (f) Output after the blind transformation, which is same as the original latent tree. }
\label{fig:CLB_example}
\end{figure}

We now describe the CLBlind algorithm, which involves two main steps. Firstly,      $\MST(V;\bD)$ is constructed  using the distance matrix $\bD$.    Secondly, we apply the blind transformation  of the Chow-Liu tree $\Blind(\MST(V;\bD))$, which proceeds as follows: 
\begin{enumerate}
\item Identify  the set of internal nodes  in $\MST(V;\bD)$.   We perform an operation for each internal node as follows:
\item \label{item:internal} For  internal node $i$, add a hidden node $h$ to the tree.  
\item \label{item:connect_edge} Connect an edge between $h$ and $i$ (which now becomes a leaf node) and also connect edges between $h$ and the neighbors of $i$ in the {\em current} tree model. 
\item Repeat steps \ref{item:internal} and \ref{item:connect_edge} until all internal nodes have been operated on. 
\end{enumerate}
See Figure~\ref{fig:CLB_example} for an illustration of CLBlind.  We use the adjective \emph{blind} to describe the transformation $\Blind(\MST(V;\bD))$ since it does not depend on the distance matrix $\bD$ but uses {\em only} the structure of the MST.   The following theorem whose proof can be found in Appendix~\ref{app:th:CLBlind} states the correctness result for CLBlind. 

\begin{THEO}\label{th:CLBlind}({\bf Correctness and Computational Complexity of CLBlind}) If $p\in \calP(\calT_{\blind})$ is a blind tree-structured graphical model Markov on $T_p$ and the matrix of distances $\bD$ is known,  then   CLBlind   outputs the true latent tree $T_p$ correctly in time $O(m^2\log m )$. \end{THEO} \vspace{0.05in} 

The first condition on $\calP(\calT_{\blind})$ that all internal nodes are hidden is not  uncommon in applications. For example, in phylogenetics, (DNA or amino acid) sequences of  extant species at the leaves are observed, while the sequences of the extinct species are hidden (corresponding to the internal nodes), and the evolutionary (phylogenetic) tree is to be reconstructed. However, the  second condition is more restrictive\footnote{The second condition on $\calP(\calT_{\blind})$ holds when the tree is (almost) homogeneous. } since it implies that each hidden node is connected to at least one observed node and that it is closer (i.e., more correlated) to one of its observed children compared to any other observed node.  
If the first constraint is satisfied but not the second, then the blind transformation $\Blind(\MST(V;\bD))$ does not overestimate the number of hidden variables in the latent tree (the proof follows from Lemma \ref{lemma:surrogate} and is omitted). 

Since the computational complexity of constructing  the $\MST$ is $O(m^2\log m)$ where $m=|V|$, and the blind transformation is at most linear in $m$, the overall computational complexity is $O(m^2\log m)$. Thus, CLBlind is a  computationally efficient procedure compared to RG, described in Section~\ref{sec:rg_exact}. 

\subsection{Chow-Liu Grouping  Algorithm}
\label{sec:clgrouping}
Even though CLBlind is computationally efficient, it only succeeds in recovering     latent trees for a restricted   subclass of minimal latent trees. In this section, we propose an efficient algorithm, called CLGrouping that reconstructs \emph{all} minimal latent trees. We also illustrate CLGrouping using an example. CLGrouping uses the properties of the MST as described in Lemma~\ref{lemma:surrogate}.

At a high-level, CLGrouping involves two distinct steps: Firstly, we construct the Chow-Liu tree $\MST(V;\bD)$  over the set of observed nodes $V$. Secondly, we apply RG or NJ to reconstruct a  latent subtree over the closed neighborhoods of every internal node in $\MST(V;\bD)$.  If RG (respectively NJ) is used, we term the algorithm CLRG (respectively CLNJ).  In the rest of the section, we only describe CLRG for concreteness since CLNJ proceeds along similar lines.  Formally,   CLRG  proceeds as follows:
\begin{enumerate}
\item \label{item:chow-liu} Construct the Chow-Liu tree $\MST(V;\bD)$ as in \eqref{eqn:TMST}. Set $T=\MST(V;\bD)$. 
\item Identify the set of internal nodes in $\MST(V;\bD)$. 
\item \label{item:cl_int} For each internal node $i$, let $\nbd[i;T]$ be its   closed neighborhood in $T$ and let $S=\RG(\nbd[i;T], \bD)$ be the output of RG with $\nbd[i;T]$ as the set of  input nodes. 
\item \label{item:cl_replace} Replace the subtree over node set $\nbd[i;T]$ in $T$ with $S$. Denote the new tree as $T$. 
\item Repeat steps \ref{item:cl_int}  and \ref{item:cl_replace} until all internal nodes have been operated on. 
\end{enumerate}
Note that the only difference between the algorithm we just described and CLNJ is Step~\ref{item:cl_int} in which the subroutine   NJ replaces RG. Also, observe in Step~\ref{item:cl_int} that RG is only applied to a small subset of nodes which have been identified in Step~\ref{item:chow-liu} as possible neighbors in the true latent tree. This reduces the computational complexity of CLRG compared to RG as seen in the following theorem whose proof is provided in Appendix~\ref{app:th:CLGrouping}. Let $|J| := |V\setminus   \Leaf(\MST(V;\bD))|<m$ be the number of internal nodes in the MST. 

\begin{THEO}\label{th:CLGrouping}({\bf Correctness and Computational Complexity of CLRG}) If $T_p\in\calT_{\ge 3}$ is a minimal latent tree and the matrix of information distances $\bD$ is available, then  CLRG outputs the true latent tree $T_p$ correctly in time $O(m^2\log m+ |J|\Delta^3(\MST(V;\bD)) )$. \end{THEO} \vspace{0.05in} 

Thus, the computational complexity of CLRG is low   when the latent tree $T_p$ has a  small maximum degree  and a small effective depth (such as the HMM) because  \eqref{eqn:surrogate:item2} implies that $\Delta(\MST(V;\bD))$ is also small. Indeed, we demonstrate in Section~\ref{sec:simulations} that there is a significant speedup compared to applying RG  over the entire observed node set $V$.

\begin{figure}[t]
\includegraphics[width=\linewidth]{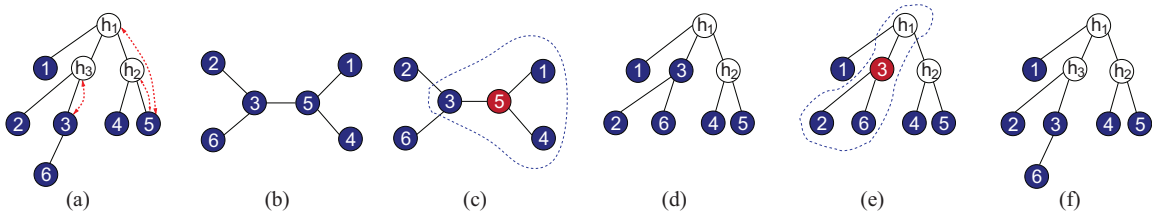} 
\caption{Illustration of CLRG. The shaded nodes are the observed nodes and the rest are hidden nodes. The dotted lines denote surrogate mappings for the hidden nodes so for example, node 3 is the surrogate of $h_3$. (a) The original latent tree, (b) The  Chow-Liu tree  (MST) over the observed nodes $V$, (c) The closed neighborhood of node $5$ is the input to RG, (d)  Output after the first RG procedure, (e) The closed neighborhood of node $3$ is the input to the second iteration of RG, (f) Output after the second RG procedure, which is same as the original latent tree. }
\label{fig:CL_example}
\end{figure}

We now illustrate CLRG using the example shown in Figure~\ref{fig:CL_example}.  The original minimal latent tree $T_p=(W,E)$ is shown in Figure~\ref{fig:CL_example}(a) with $W=\{1,2,\ldots,6, h_1,h_2,h_3\}$. The  set of  observed nodes is $V=\{1,\ldots,6\}$ and the set of hidden nodes is $H=\{h_1,h_2,h_3\}$. The Chow-Liu tree $T_{\CL}=\MST(V; \bD)$ formed using the information distance matrix $\bD$  is shown in Figure~\ref{fig:CL_example}(b). Since nodes $3$ and $5$ are the only internal nodes in $\MST(V; \bD)$, two RG operations will be executed on the closed neighborhoods of each of these two nodes. In the first iteration, the closed neighborhood of node $5$ is the input to RG. This is shown in Figure~\ref{fig:CL_example}(c) where $\nbd[4;\MST(V;\bD)]=\{1,3,4,5\}$, which is then replaced by the output of RG to obtain the tree shown in Figure~\ref{fig:CL_example}(d).  In the next iteration, RG is applied to the closed neighborhood of node 3 in the current tree $\nbd[3;T]=\{2,3,6,h_1\}$   as shown in Figure~\ref{fig:CL_example}(e).  Note that $\nbd[3;T]$ includes $h_1\in H$, which was introduced by RG in the previous iteration. This closed neighborhood is then replaced by the output of the second RG operation and the  original latent tree $T_p$ is obtained as shown in Figure~\ref{fig:CL_example}(f). 

Observe  that the trees obtained at each iteration of CLRG are  related to the original latent tree in terms of edge-contraction operations \citep{Rob81}, which were defined in Section~\ref{sec:cl_latent}. For example, the Chow-Liu tree in Figure~\ref{fig:CL_example}(b) is obtained from the latent tree $T_p$ in Figure~\ref{fig:CL_example}(a) by sequentially contracting all edges connecting an observed node to its inverse surrogate set (cf.\ Lemma~\ref{lemma:surrogate}(ii)). Upon performing an iteration of RG, these contraction operations are inverted and hidden nodes are introduced. For example, in Figure~\ref{fig:CL_example}(d), the hidden nodes   $h_1,h_2$ are introduced after performing RG on the closed neighborhood of node $4$ on $\MST(V;\bD)$. These newly introduced hidden nodes in fact, turn out to be the inverse surrogate set of node $4$, i.e., $\Sg^{-1}(5)= \{5,h_1,h_2\}$.   This is not merely a coincidence and we prove in Appendix~\ref{app:th:CLGrouping} that at each iteration, the set of hidden nodes introduced corresponds to the inverse surrogate set of the internal node.


%
%

We conclude this section by emphasizing that CLGrouping (i.e., CLRG or CLNJ) has two primary advantages. Firstly, as demonstrated in Theorem~\ref{th:CLGrouping},  the structure of  all minimal tree-structured graphical models  can be recovered by CLGrouping in contrast to   CLBlind. Secondly, it typically has much lower computational complexity   compared to RG. 

%

\subsection{Extension to General Discrete Models}
\label{sec:general_discrete}

For general   (i.e., not symmetric)  discrete models, the mutual information $I(X_i\,;\, X_j)$ is in general   not monotonic in the  information distance $d_{ij}$, defined   in~\eqref{eq:dist_discrete}.\footnote{The mutual information, however, is monotonic in $d_{ij} $ for asymmetric binary discrete models.} As a result, Lemma~\ref{lemma:dist} does not hold, i.e., the Chow-Liu tree  ${T}_{\ML}$ is not necessarily  the same as $\MST(V;\bD)$.  However, Lemma~\ref{lemma:surrogate} does hold for all minimal latent tree models.  Therefore, for general discrete models, we compute $\MST(V;\bD)$ (instead of the Chow-Liu tree ${T}_{\ML}$ with edge weights $I(X_i\, ;\, X_j)$), and apply RG or NJ to each internal node and its neighbors.  This algorithm guarantees that the structure learned using CLGrouping is the same as $T_p$ if  the distance matrix   $\bD$ is available.


\section{Sample-Based Algorithms for Learning  Latent Tree Structures}
\label{sec:rg_estimate}
In Sections~\ref{sec:rg_exact} and~\ref{sec:chowLiu_exact}, we designed algorithms for the exact reconstruction of latent trees  assuming that $p_V$ is a tree-decomposable distribution and the matrix of information distances $\bD$ is available. In most (if not all) machine learning problems, the pairwise distributions $p(x_i,x_j)$ are unavailable. Consequently, $\bD$  is also unavailable so RG, NJ and CLGrouping as stated in Sections~\ref{sec:rg_exact} and~\ref{sec:chowLiu_exact} are not directly applicable. In this section, we consider extending RG, NJ and CLGrouping to the case when only samples $\bx_V^n$ are available. We show how to modify the previously proposed algorithms to accommodate ML estimated distances  and we also provide sample complexity results for {\em relaxed} versions of  RG and CLGrouping. 

\subsubsection*{ML Estimation of Information Distances}
The canonical method for deterministic parameter estimation is via maximum-likelihood (ML) \citep{Serfling}. We focus on Gaussian and symmetric discrete distributions in this section. The generalization to general discrete models is straightforward. For Gaussians graphical models, we use ML   to estimate the entries of the covariance matrix,\footnote{Recall that we assume that the mean of the true random vector $\bX$ is known and equals to the zero vector so we do not need to subtract the empirical mean in \eqref{eqn:emp_cov}. } i.e.,
\begin{equation}
\hSigma_{ij}=\frac{1}{n}\sum_{k=1}^n x_{i}^{(k)} x_{j}^{(k)},\qquad\forall\, i,j\in V. \label{eqn:emp_cov}
\end{equation}
The ML estimate of the correlation coefficient is defined as  $\hrho_{ij} :=  \hSigma_{ij}/ (\hSigma_{ii} \hSigma_{jj})^{1/2}$. The estimated information distance is then given by the analogue of~\eqref{eq:dist_gaussian}, i.e., $\hd_{ij} = -\log|\hrho_{ij}|$.  For symmetric discrete distributions, we estimate the crossover probability $\theta_{ij}$ via ML as\footnote{We use $\mathbb{I}\{ \cdot\}$ to  denote the indicator function. }
\begin{equation}
\htheta_{ij} =\frac{1}{n}\sum_{k=1}^n \mathbb{I} \big\{ x_i^{(k)} \ne  x_j^{(k)} \big\},\qquad\forall\, i,j\in V. 
\end{equation}
The estimated information distance is given by the analogue of \eqref{eq:dist_symdiscrete}, i.e.,  $\hd_{ij}=-(K-1)\log (1-K\htheta_{ij})$. For both classes of models, it can easily be verified from the  Central Limit Theorem and continuity arguments \citep{Serfling} that $ \hd_{ij} -d_{ij} =O_p(n^{-1/2})$, where $n$ is the number of samples. This means that the estimates of the information distances are consistent with      rate of convergence being $n^{-1/2}$.  The  $m\times m$ matrix of estimated information distances is denoted as $\hbD=[\hd_{ij}]$. 

\subsection{Relaxed Recursive Grouping (RG) Given Samples}
\label{subsec:rg_estimate}
We now show how to relax the canonical RG algorithm described in Section~\ref{sec:rg_exact} to   handle the case when only $\hbD$ is available. Recall that RG calls  the $\TestNodeRelationships$ procedure recursively to ascertain child-parent and sibling relationships via equality tests  $\Phi_{ijk}=d_{ik}-d_{jk}$ (cf.\ Section~\ref{sec:testing_node}). These equality constraints are, in general,   not satisfied with the estimated differences $\hPhi_{ijk}:= \hd_{ik}-\hd_{jk}$, which are computed based on the estimated distance in $\hbD$.  Besides, not all estimated distances are equally accurate. Longer distance estimates (i.e., lower correlation estimates) are less accurate for a given number of samples.\footnote{In fact, by using  a large deviation result in \citet[Theorem~1]{Shen}, we can formally show that a larger number of samples is    required to get a good approximation of $\rho_{ik}$ if it is small compared to when $\rho_{ik}$ is large.  }  As such, not all estimated distances can be used for testing inter-node relationships reliably.   These observations motivate the following three modifications to the RG algorithm: 
\begin{enumerate}
\item Consider using a smaller subset of nodes to test whether $\hPhi_{ijk}$ is constant (across $k$).
\item Apply a threshold (inequality) test to the  $\hPhi_{ijk}$ values.
\item Improve on the robustness of the estimated distances $\hd_{ih}$  in \eqref{eq:dist_childparent_exact} and \eqref{eq:dist_hiddenother_exact} by averaging. 
\end{enumerate}
We now describe each of these modifications in greater detail. Firstly, in  the relaxed RG algorithm, we only compute $\hPhi_{ijk}$ for those estimated distances $\hd_{ij}$,  $\hd_{ik}$ and $\hd_{jk}$ that  are below a prescribed threshold  $\tau>0$ since longer distance estimates are unreliable. As such, for each pair of nodes $(i,j)$ such that $\hd_{ij} < \tau$, associate the set 
\begin{equation}
\calK_{ij} := \left\{k \in  V \backslash \{i,j\}: \max\{\hd_{ik}  , \hd_{jk}\} < \tau\right\}. \label{eqn:calKij}
\end{equation}
This is the subset of nodes in $V$ whose estimated distances to $i$ and $j$ are less than $\tau$. Compute $\hPhi_{ijk}$ for all $k\in \calK_{ij}$ only. 

Secondly, instead of using equality tests in $\TestNodeRelationships$ to determine the relationship between nodes $i$ and $j$, we relax this test and consider the statistic
\begin{equation}
\hdiss_{ij} := \max_{k \in \calK_{ij} } \widehat{\Phi}_{ijk} - \min_{k \in \calK_{ij} } \widehat{\Phi}_{ijk} \label{eqn:phi_diff}
\end{equation}
Intuitively, if $\hdiss_{ij}$ in~\eqref{eqn:phi_diff} is close to zero, then  nodes $i$ and $j$ are likely to be in the same family.  Thus, declare that  nodes $i,j\in V$ are in the same sibling group if 
\begin{equation}
\hdiss_{ij}<\epsilon,  \label{eqn:thres_eps} 
\end{equation}
for another threshold $\epsilon>0$. Similarly, an observed node $k$ is identified as a parent node if $|\hd_{ik} + \hd_{kj} - \hd_{ij}| < \epsilon$ for all $i$ and $j$ in the sibling group.    



Thirdly, in order to further improve on the quality of the distance estimate  $\hd_{ih}$ of a newly introduced hidden node to observed nodes, we compute $\hd_{ih}$ using (\ref{eq:dist_childparent_exact}) with different pairs of $j \in \calC(h)$ and $k \in \calK_{ij}$, and take the average as follows:
\begin{equation}
\hd_{ih} = \frac{1}{2 (|\calC(h)|-1) }\left( \,\sum_{j \in \calC(h)} \hd_{ij} + \frac{1}{|\calK_{ij} |} \sum_{k \in \calK_{ij}} \hPhi_{ijk} \right).
\label{eq:dist_childparent}
\end{equation}
\noindent 
Similarly, for any other node $k \notin \calC(h)$, we compute $\hd_{kh}$ using all child nodes in $\calC(h)$ and   $\calC(k)$ (if $\calC(k)\ne \emptyset$)    as follows:
\begin{equation}
\hd_{kh} = \left\{
\begin{array}{ll}
\frac{1}{|\calC(h)|} \sum_{i \in \calC(h)} (\hd_{ik}- \hd_{ih}), & \mathrm{if}~k \in V,\\
\frac{1}{|\calC(h)| |\calC(k)|} \sum_{(i,j) \in \calC(h)\times \calC(k)} ( \hd_{ij} - \hd_{ih} - \hd_{jk}), &\mathrm{otherwise.}
\end{array}
\right.
\label{eq:dist_hiddenother}
\end{equation}
It is easy to verify that if $\hd_{ih}$ and $\hd_{kh}$   are equal to $d_{ih}$ and $d_{kh}$ respectively, then~(\ref{eq:dist_childparent}) and~(\ref{eq:dist_hiddenother}) reduce to~(\ref{eq:dist_childparent_exact}) and~(\ref{eq:dist_hiddenother_exact})  respectively.

The following theorem shows that relaxed RG is consistent, and with appropriately chosen thresholds $\epsilon$ and $\tau$, it has the sample complexity logarithmic in the number of observed variables.  The proof follows from standard Chernoff bounds and is provided in Appendix~\ref{proof:rg_sample}. 

{\THEO {\bf (Consistency and Sample Complexity of Relaxed RG)}  \label{th:rg_sample}
(i) Relaxed RG is structurally consistent for all $T_p \in \calT_{\ge 3}$.  In addition, it is risk consistent for Gaussian and symmetric discrete distributions.
(ii) Assume that the effective depth is $\delta(T_p; V)=O(1)$ (i.e., constant in $m$) and relaxed RG is used to reconstruct the  tree given $\hbD$. For every $\eta>0$, there exists thresholds $\epsilon, \tau>0$ such that if 
\begin{equation}
n>C \, \log (m/\sqrt[3]{\eta}) \label{eqn:sample_complex_rg}
\end{equation}
for some constant $C >0$,  the error probability for structure reconstruction in~\eqref{eqn:structural_consistency} is bounded above by $\eta$. If, in addition,  $p$ is a Gaussian or symmetric discrete distribution  and  $n>C'\log (m/\sqrt[3]{\eta})$, the error probability for distribution reconstruction in \eqref{eqn:consistency} is also bounded above by $\eta$.  Thus, the sample complexity of relaxed RG, which is the number of samples required to achieve a desired level of accuracy, is logarithmic in $m$, the number of observed variables.
\label{thm:rg_sample} }
\vspace{0.05in}

As we observe from \eqref{eqn:sample_complex_rg}, the sample complexity for RG is logarithmic in $m$ for shallow trees (i.e., trees where the effective depth is constant). This is in contrast to NJ where the sample complexity is super-polynomial in the number of observed nodes for the HMM \citep{StJohn03,lacey06}. 

\subsubsection*{RG with $k$-means Clustering}
In practice, if the number of samples is limited, the distance estimates $\hd_{ij}$  are noisy and it is difficult to select the threshold $\epsilon$ in Theorem~\ref{th:rg_sample} to identify sibling nodes reliably. In our experiments, we employ a modified version of the $k$-means clustering algorithm to cluster a set of nodes with small $\hdiss_{ij}$, defined in \eqref{eqn:phi_diff}, as a family. Recall that we test each $\hdiss_{ij}$ locally with a fixed threshold $\epsilon$  in \eqref{eqn:thres_eps}. In contrast, the $k$-means   algorithm provides a {\em global} scheme and circumvents the need to select the threshold $\epsilon$. 
We adopt the  \emph{silhouette method} \citep{rousseeuw87} with dissimilarity measure  $\hdiss_{ij}$ to select optimal the number of clusters~$k$.

\subsection{Relaxed Neighbor-Joining Given Samples} \label{subsec:nj_estimate}

In this section, we describe how NJ can be relaxed when the true distances are unavailable.  We relax the NJ algorithm by using ML estimates of the distances $\hd_{ij}$  in place of unavailable distances $d_{ij}$. NJ typically assume that all observed nodes are at the leaves of the latent tree, so after learning the latent tree, we perform the following post-processing step: If there exists an edge $(i,h)\in W\times H$ with $\hd_{ih}<\epsilon'$ (for a  given  threshold $\epsilon'>0$), then $(i,h)$ is contracted to a single node whose label is $i$.
The sample complexity of NJ  is known to be $O(\exp(\diam(T_p))\log m)$ \citep{StJohn03} and thus does not scale well when the latent tree $T_p$ has a large diameter. 
Comparisons between the sample complexities of other  closely related latent tree learning algorithms are  discussed in \cite{Atteson:99Algo,erdos99,Csu00} and \cite{StJohn03}. 




\subsection{Relaxed CLGrouping Given Samples} \label{subsec:clgroup_estimate}
In this section, we discuss how to modify CLGrouping (CLRG and CLNG) when we only have access to the estimated information distance $\hbD$. The relaxed version of CLGrouping differs from CLGrouping in two main aspects. Firstly, we replace the edge weights in the construction of the MST in \eqref{eqn:TMST} with the estimated information distances $\hd_{ij}$, i.e., 
\begin{equation}
\hT_{\CL} =\MST( V;\hbD) := \argmin_{ T\in \calT(V)} \,\, \sum_{(i,j) \in T}\,\, \hd_{ij}.  \label{eqn:hTMST}
\end{equation}
The procedure in \eqref{eqn:hTMST} can be shown to be equivalent to the learning of the ML tree structure given samples $\bx_V^n$ if $p_V$  is a Gaussian or symmetric discrete distribution.\footnote{This follows from the observation that the ML search for the optimal structure is equivalent to the KL-divergence minimization problem in \eqref{eqn:q_tree} with $p_V$ replaced by $\hp_V$, the empirical distribution of $\bx^n_V$. }   It has also been shown that the error probability of structure learning    $\Pr(\hT_{\CL}\ne T_{\CL})$ converges to zero exponentially fast in the number of samples $n$ \citep{Tan09,tan10}.  Secondly, for CLRG (respectively CLNJ), we replace RG  (respectively NJ) with the relaxed version of RG (respectively NJ). The sample complexity result of CLRG (and its proof) is  similar to Theorem~\ref{th:rg_sample} and 
the proof is provided in Appendix~\ref{proof:cl_sample}.

{\THEO {\bf (Consistency and Sample Complexity of Relaxed CLRG)}  \label{th:cl_sample}
(i) Relaxed CLRG is structurally consistent for all $T_p \in \calT_{\ge 3}$.  In addition, it is risk consistent for Gaussian and symmetric discrete distributions.
(ii) Assume that   the effective depth is $\delta(T_p; V)=O(1)$ (i.e., constant in $m$). 
Then  the sample complexity of relaxed CLRG is logarithmic in $m$.
}
\vspace{0.05in}

After CLRG has been completed, as a final post-processing step, if we find that  there exists an  estimated distance  $\hd_{ih}$ on an edge with $i \in W$ and $h \in H$ in the learned model which is smaller than some $\epsilon'>0$ (which we specify in our experiments), then edge $(i,h)$ is {\em contracted} to the single node $i$. This is similar to relaxed NJ and serves to contract all strong edges in the learned  model. 

\subsection{Regularized CLGrouping for Learning Latent Tree Approximations} \label{subsec:regCLG}

For many practical applications, it is of interest to learn a latent tree that \emph{approximates} the given empirical distribution.  
In general, introducing more hidden variables enables better fitting to the empirical distribution, but it increases the model complexity and may lead to overfitting.  The Bayesian Information Criterion \citep{schwarz78} provides a trade-off between model fitting and model complexity, and is defined as follows:
\begin{equation}
\mathrm{BIC}(\hT) = \log p(\bx^n_{V} ; \hT) - \frac{\kappa(\hT)}{2} \log n
\label{eq:bic}
\end{equation}
\noindent where $\hT$ is a latent tree structure and $\kappa(\hT)$ is the number of free parameters, which grows linearly with the number of hidden variables because $\hT$ is a tree.  
Here, we describe {\em regularized CLGrouping}, in which we use the BIC in \eqref{eq:bic} to specify a stopping criterion on the number of hidden variables added. 

For each internal node and its neighbors in the Chow-Liu tree, we use relaxed NJ or RG to learn a latent subtree.  Unlike in regular CLGrouping, before we integrate this subtree into our model, we compute its BIC score.  Computing the BIC score requires estimating the maximum likelihood parameters for the models, so for general discrete distributions, we run the EM algorithm on the subtree to estimate the parameters.\footnote{Note that for Gaussian and symmetric discrete distributions, the model parameters can be recovered from information distances directly using (\ref{eq:dist_gaussian}) or (\ref{eq:dist_symdiscrete}).}  After we compute the BIC scores for all subtrees corresponding to all internal nodes in the Chow-Liu tree, we choose the subtree that results in the highest BIC score and incorporate that subtree into the current tree model.  

The BIC score can be computed efficiently on a tree model with a few hidden variables.  Thus, for computational efficiency, each time a set of hidden nodes is added to the model, we generate samples of hidden nodes conditioned on the samples of observed nodes, and use these augmented samples to compute the BIC score approximately when we evaluate the next subtree to be integrated in the model.

If none of the subtrees increases the BIC score (i.e., the current tree has the highest BIC score), the procedure stops and outputs the estimated latent tree.  Alternatively, if we wish to learn a latent tree with a given number of hidden nodes, we can used the BIC-based procedure mentioned in the previous paragraph to learn subtrees until the desired number of hidden nodes is introduced.  Depending on whether we use NJ or RG as the subroutine, we denote the specific regularized CLGrouping algorithm as {\em regCLNJ} or {\em regCLRG}.

This approach of using an approximation of the BIC score has been commonly used to learn a graphical model with hidden variables \citep{elidan05,zhang04}.  However, for these algorithms, the BIC score needs to be evaluated for a large subset of nodes, whereas in CLGrouping, the Chow-Liu tree among observed variables prunes out many subsets, so we need to evaluate BIC scores only for a small number of candidate subsets (the number of internal nodes in the Chow-Liu tree).

\section{Experimental Results}
\label{sec:simulations}

In this section, we compare the performances of various latent tree learning algorithms.  We first show simulation results on synthetic datasets with known latent tree structures to demonstrate the consistency of our algorithms. We also analyze the performance of these algorithms when we change the underlying latent tree structures.  Then, we show that our algorithms can approximate  arbitrary multivariate probability distributions with latent trees by applying them to two real-world datasets, a monthly stock returns example and the 20 newsgroups dataset.  

\begin{figure}[t]
\centerline{\includegraphics[width=\linewidth]{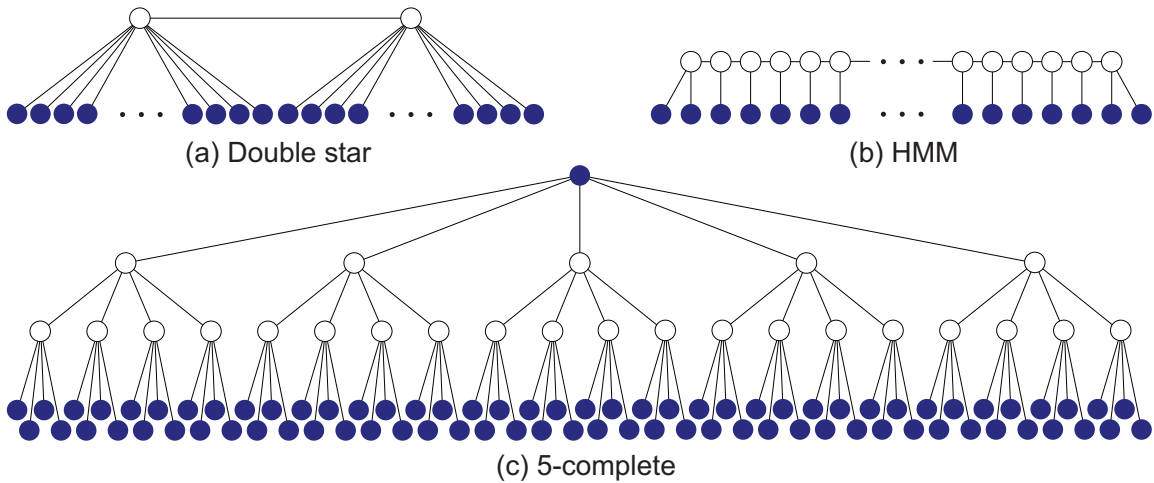}}
\caption{Latent tree structures used in our simulations.}
\label{fig:synthetic_trees}
\end{figure}

\subsection{Simulations using Synthetic Datasets}

In order to analyze the performances of different tree reconstruction algorithms, we generate samples from known latent tree structures with varying sample sizes and apply reconstruction algorithms.  We compare the neighbor-joining method (NJ) \citep{Sai87} with recursive grouping (RG), Chow-Liu Neighbor Joining (CLNJ), and Chow-Liu Recursive Grouping (CLRG).  Since the algorithms are given only samples of observed variables, we use the sample-based algorithms described in Section \ref{sec:rg_estimate}.  For all our experiments, we use the same edge-contraction threshold $\epsilon' = -\log 0.9$ (see Sections \ref{subsec:nj_estimate} and \ref{subsec:clgroup_estimate}), and set $\tau$ in Section \ref{subsec:rg_estimate} to grow logarithmically with the number of samples.

Figure~\ref{fig:synthetic_trees} shows the three latent tree structures used in our simulations. The double-star has 2 hidden   and 80 observed nodes, the HMM has 78 hidden and 80 observed nodes, and the 5-complete tree has 25 hidden and 81 observed nodes including the root node. For simplicity, we present simulation results only on Gaussian models but note that the behavior on discrete models is similar.  All correlation coefficients on the edges $\rho_{ij}$ were   independently   drawn from a uniform distribution supported on $[0.2, 0.8]$.  The performance of each method is measured by averaging over 200 independent runs with different parameters. We use the following performance metrics to quantify  the performance of each algorithm in Figure~\ref{fig:results_synthetic}:

\begin{figure}[t]
\centerline{\includegraphics[width=1\linewidth]{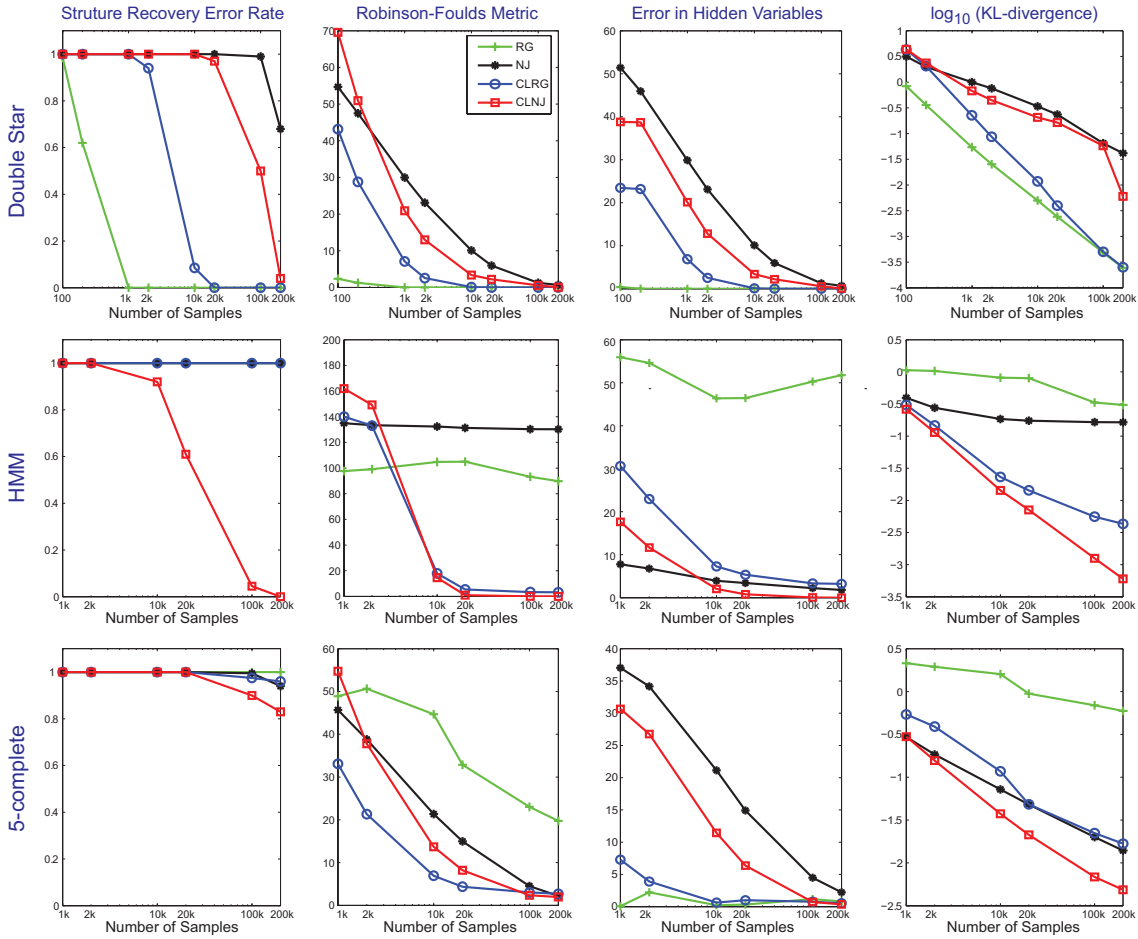}}
\caption{Performance of RG, NJ, CLRG, and CLNJ for the  latent trees shown in Figure~\ref{fig:synthetic_trees}.} 
\label{fig:results_synthetic}
\end{figure}


\begin{enumerate}
\item[(i)] {\bf Structure recovery error rate}: This is the proportion of times that the proposed algorithm fails to recover the true latent tree structure. Note that this is a very strict measure since even a single wrong hidden node or misplaced edge results in an error for the entire structure. 
\item[(ii)] {\bf Robinson Foulds metric} \citep{Rob81}: This popular  phylogenetic tree-distortion metric computes  the number of graph transformations (edge contraction or expansion) needed to be applied to the estimated graph in order to get the correct structure.  This metric quantifies the   difference in the structures of the estimated and true models. 
\item[(iii)] {\bf Error in the number of hidden variables}: We compute the average number of hidden variables introduced by each method and plot the absolute difference between the average estimated hidden variables and the number of hidden variables in the true structure.
\item[(iv)] {\bf KL-divergence} $D(p_V \, || \, \hp_V^n)$: This is a measure of the distance between the estimated and the true models over the set of  observed nodes $V$.\footnote{Note that this is not the same quantity as in~\eqref{eqn:consistency} because if the number of  hidden variables is estimated incorrectly, $D(p \,||\, \hp^n)$ is infinite so we plot $D(p_V \, ||\, \hp_V^n)$ instead.  However, for Gaussian and symmetric discrete distributions, $D(p \, ||\, \hp^n)$ converges to zero in probability since the number of hidden variables is estimated correctly asymptotically. } 
\end{enumerate}
We first note that from the structural error rate plots that the double star is the easiest structure to recover and the 5-complete tree is the hardest.  In general, given the same number of observed variables, a latent tree with more hidden variables or larger effective depth (see Section \ref{sec:model}) is more difficult to recover.  

For the double star, RG clearly outperforms all other methods.  With only 1,000 samples, it recovers the true structure exactly in all 200 runs.  
On the other hand, CLGrouping performs significantly better than RG for the HMM.  There are two reasons for such performance differences.  Firstly, for Gaussian distributions, it was shown \citep{tan10} that given the same number of variables and their samples, the Chow-Liu algorithm is most accurate for a chain and least accurate for a star.  Since the Chow-Liu tree of a latent double star graph is close to a star, and the Chow-Liu tree of a latent HMM is close to a chain, the Chow-Liu tree tend to be more accurate for the HMM than for the double star.  Secondly, the internal nodes in the Chow-Liu tree of the HMM tend to have small degrees, so we can apply RG or NJ to a very small neighborhood, which results in a significant improvement in both accuracy and computational complexity.


Note that NJ is particularly poor at recovering the HMM structure.  In fact, it has been shown that even if  the number of samples  grows polynomially with  the number of observed variables  (i.e., $n = O(m^{B})$ for any $B>0$), it is insufficient for NJ to recover HMM structures  \citep{lacey06}. The 5-complete tree has two layers of hidden nodes, making it very difficult to recover the exact structure using any method.  CLNJ has the best structure recovery error rate and KL divergence, while CLRG has the smallest Robinson-Foulds metric.

\begin{table}[t]
\centering
\begin{tabular}{|c|c|c|c|c|}
\hline
  & RG  & NJ & CLRG & CLNJ  \\ \hline
HMM   &  10.16  &  0.02  &  0.10  &  0.05 \\ \hline
5-complete  & 7.91  &  0.02  &  0.26  &  0.06\\ \hline 
Double star  & 1.43 &   0.01 &   0.76  &  0.20 \\ \hline
\end{tabular}
\caption{Average running time of each algorithm in seconds.}
\label{table:runningtimes}
\end{table}

Table~\ref{table:runningtimes} shows the running time of each algorithm averaged over 200 runs and all sample sizes.  All algorithms are implemented in MATLAB.  As expected, we observe that CLRG is significantly faster than RG for HMM and 5-complete graphs.  NJ is fastest, but CLNJ is also very efficient and leads to much more accurate reconstruction of latent trees. 

Based on the simulation results, we conclude that for a latent tree with a few hidden variables, RG is most accurate, and for a latent tree with a large diameter, CLNJ performs the best.  A latent tree with multiple layers of hidden variables is more difficult to recover correctly using any method, and CLNJ and CLRG outperform NJ and RG.

\subsection{Monthly Stock Returns}

\begin{figure}[t]
\centerline{\includegraphics[width=0.5\linewidth]{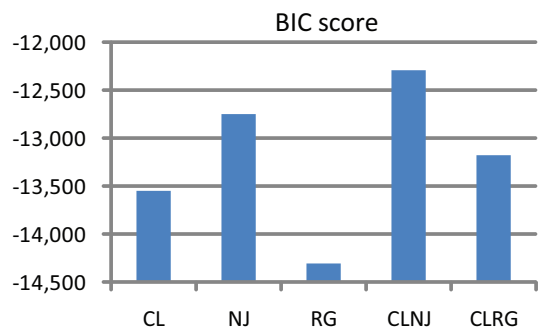}}
\caption{Plot of BIC scores for the monthly stock returns example.}
\label{fig:sp100_bic}
\end{figure}

\begin{figure}[p]
\centerline{\includegraphics[width=0.8\linewidth]{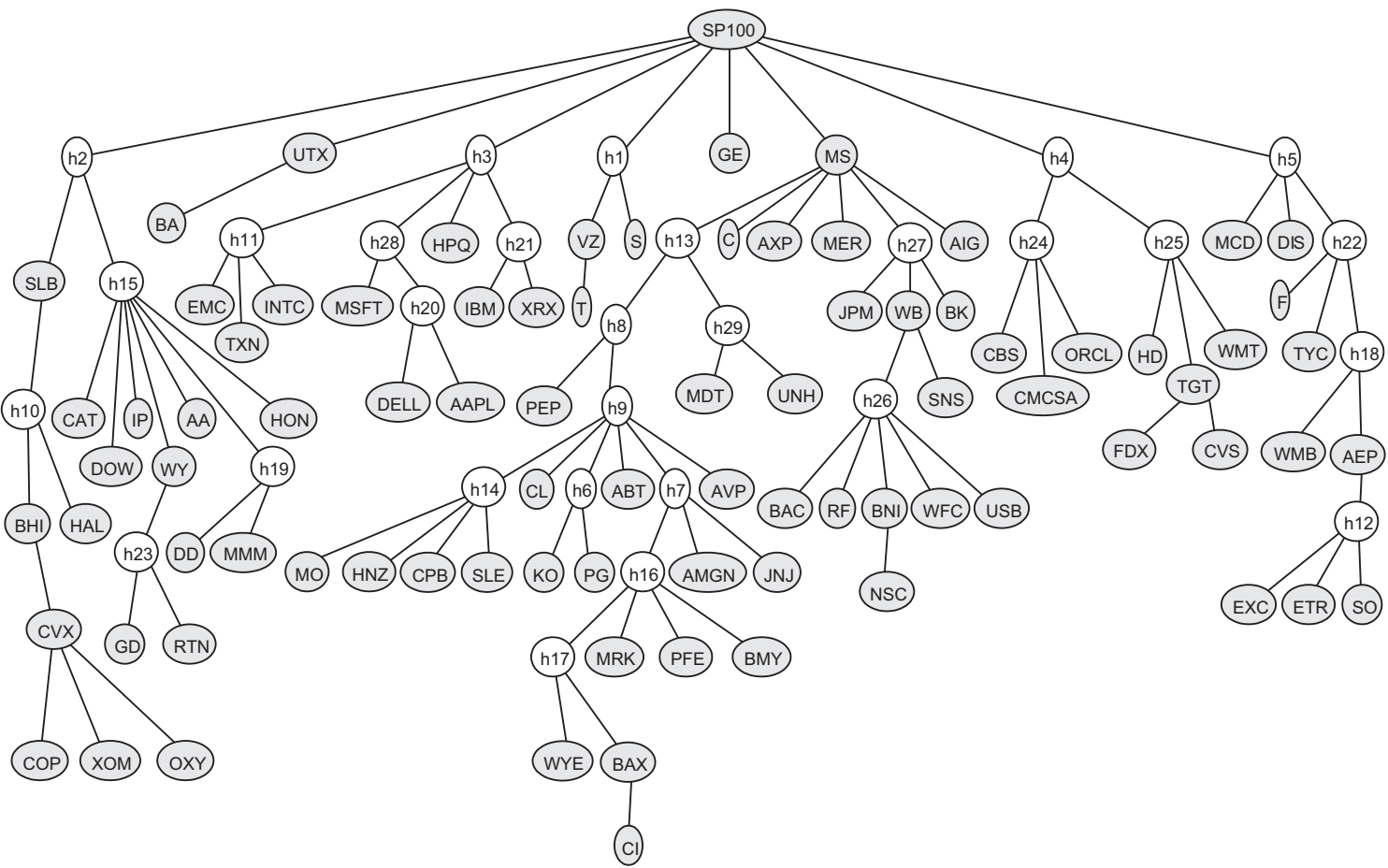}}
\caption{Tree structure learned from monthly stock returns using CLNJ.}
\label{fig:SP100_CLNJ_tree}
\end{figure}

We apply our latent tree learning algorithms to model the dependency structure of monthly stock returns of $84$ companies in the S\&P $100$ stock index.\footnote{We disregard $16$ companies that have been listed on S\&P $100$ only after $1990$.}  We use the samples of the monthly returns from $1990$ to $2007$.  As shown in Table~\ref{table:sp100_table} and Figure~\ref{fig:sp100_bic},  CLNJ achieves the highest log-likelihood and BIC scores.  NJ introduces more hidden variables than CLNJ and has lower log-likelihoods, which implies that starting from a Chow-Liu tree helps to get a better latent tree approximation.  Figure~\ref{fig:SP100_CLNJ_tree} shows the latent tree structure learned using the CLNJ method. Each observed node is labeled with the ticker of the company. Note that related companies are closely located on the tree.  Many hidden nodes can be interpreted as industries or divisions.  For example, h1 has Verizon, Sprint, and T-mobile as descendants, and can be interpreted as the telecom industry, and h3 correspond to the technology division with companies such as Microsoft, Apple, and IBM as descendants.  Nodes h26 and h27 group commercial banks together, and h25 has all retail stores as child nodes. 

\begin{table}[t]
\centering
\begin{tabular}{|c|c|c|c|c|c|}
\hline
  & Log-Likelihood & BIC & \# Hidden & \# Parameters & Time (secs) \\ \hline
CL &  -13,321   &  -13,547  &  0 &  84 & 0.15 \\ \hline
 NJ & -12,400 & -12,747  &  45 &  129 & {\bf 0.02} \\ \hline \hline
RG & -14,042 &  -14,300 &   12  &  96 & 21.15 \\ \hline
CLNJ & {\bf -11,990} &  {\bf -12,294} &   29  &  113 & 0.24 \\ \hline
CLRG & -12,879 &  -13,174 &   26  &  110 & 0.40 \\ \hline
\end{tabular}
\caption{Comparison of the log-likelihood, BIC, number of hidden variables introduced, number of parameters, and running time for the monthly stock returns example.}
\label{table:sp100_table}
\end{table}

\subsection{20 Newsgroups with 100 Words}

For our last experiment, we apply our latent tree learning algorithms to the 20 Newsgroups dataset with 100 words.\footnote{\url{http://cs.nyu.edu/~roweis/data/20news_w100.mat}}  The dataset consists of 16,242 binary samples of 100 words, indicating whether each word appears in each posting or not.  In addition to the Chow-Liu tree (CL), NJ, RG, CLNJ, and CLRG, we also compare the performances with the regCLNJ and regCLRG (described in Section \ref{subsec:regCLG}), the latent cluster model (LCM) \citep{lazarsfeld68}, and BIN, which is a greedy algorithm for learning latent trees \citep{harmeling10}.  

\begin{figure}[t]
\centerline{\includegraphics[width=0.6\linewidth]{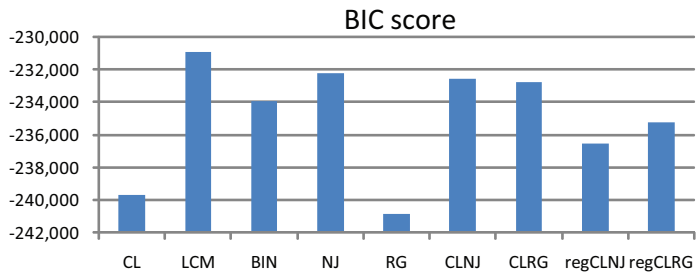}}
\caption{The BIC scores of various algorithms on the newsgroup set.}
\label{fig:newsgroup_bic_all}
\end{figure}

Table~\ref{table:newsgroup_table_all} shows the performance of different algorithms, and Figure~\ref{fig:newsgroup_bic_all} plots the BIC score.  We use the MATLAB code (a small part of it is implemented in C) provided by \cite{harmeling10}\footnote{\url{http://people.kyb.tuebingen.mpg.de/harmeling/code/ltt-1.3.tar}} to run LCM and BIN.  Note that although LCM has only one hidden node, the hidden node has 16 states, resulting in many parameters.  We also tried to run the algorithm by \cite{Che08}, but their JAVA implementation on this dataset did not complete even  after several days. For NJ, RG, CLNJ, and CLRG, we learned the structures using only information distances (defined in (\ref{eq:dist_discrete})) and then used the EM algorithm to fit the parameters.  For regCLNJ and regCLRG, the model parameters are learned during the structure learning procedure by running the EM algorithm locally, and once the structure learning is over, we refine the parameters by running the EM algorithm for the entire latent tree.  All methods are implemented in MATLAB except the E-step of the EM algorithm, which is implemented in C++.

\begin{figure}[t]
\centerline{\includegraphics[width=0.8\linewidth]{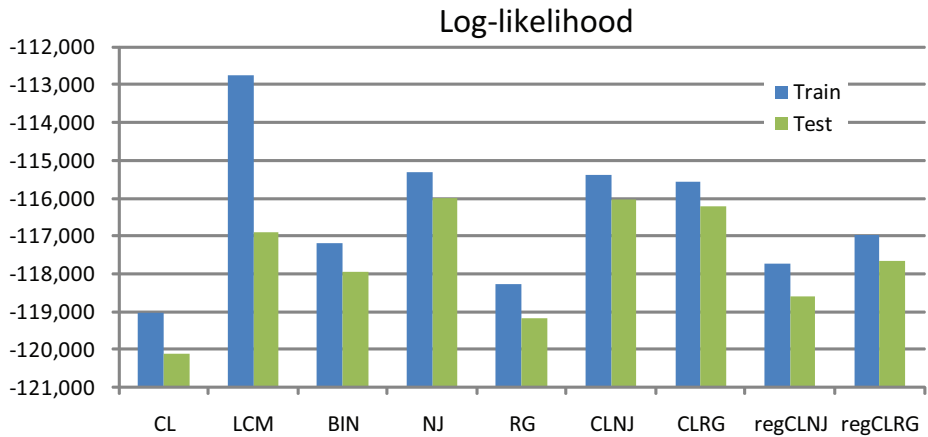}}
\caption{Train and test log-likelihood scores of various algorithms on the newsgroup dataset with a train/test split.}
\label{fig:newsgroup_ll_trte}
\end{figure}

Despite having many parameters, the models learned via LCM have the best BIC score.  However, it does not reveal any interesting structure and is computationally more expensive to learn.  In addition, it may result in overfitting.  In order to show this, we split the dataset randomly and use half as the training set and the other half as the test set.  Table~\ref{table:newsgroup_table_trte} shows the performance of applying the latent trees learned from the training set to the test set, and Figure~\ref{fig:newsgroup_ll_trte} shows the log-likelihood on the training and the test sets.  For LCM, the test log-likelihood drops significantly compared to the training log-likelihood, indicating that LCM is overfitting the training data. NJ, CLNJ, and CLRG achieve high log-likelihood scores on the test set.  Although regCLNJ and regCLRG do not result in a better BIC score, they introduce fewer hidden variables, which is desirable if we wish to learn a latent tree with small computational complexity, or if we wish to discover a few hidden variables that are meaningful in explaining the dependencies of observed variables.  

\begin{figure}[p]
\centerline{\includegraphics[width=\linewidth]{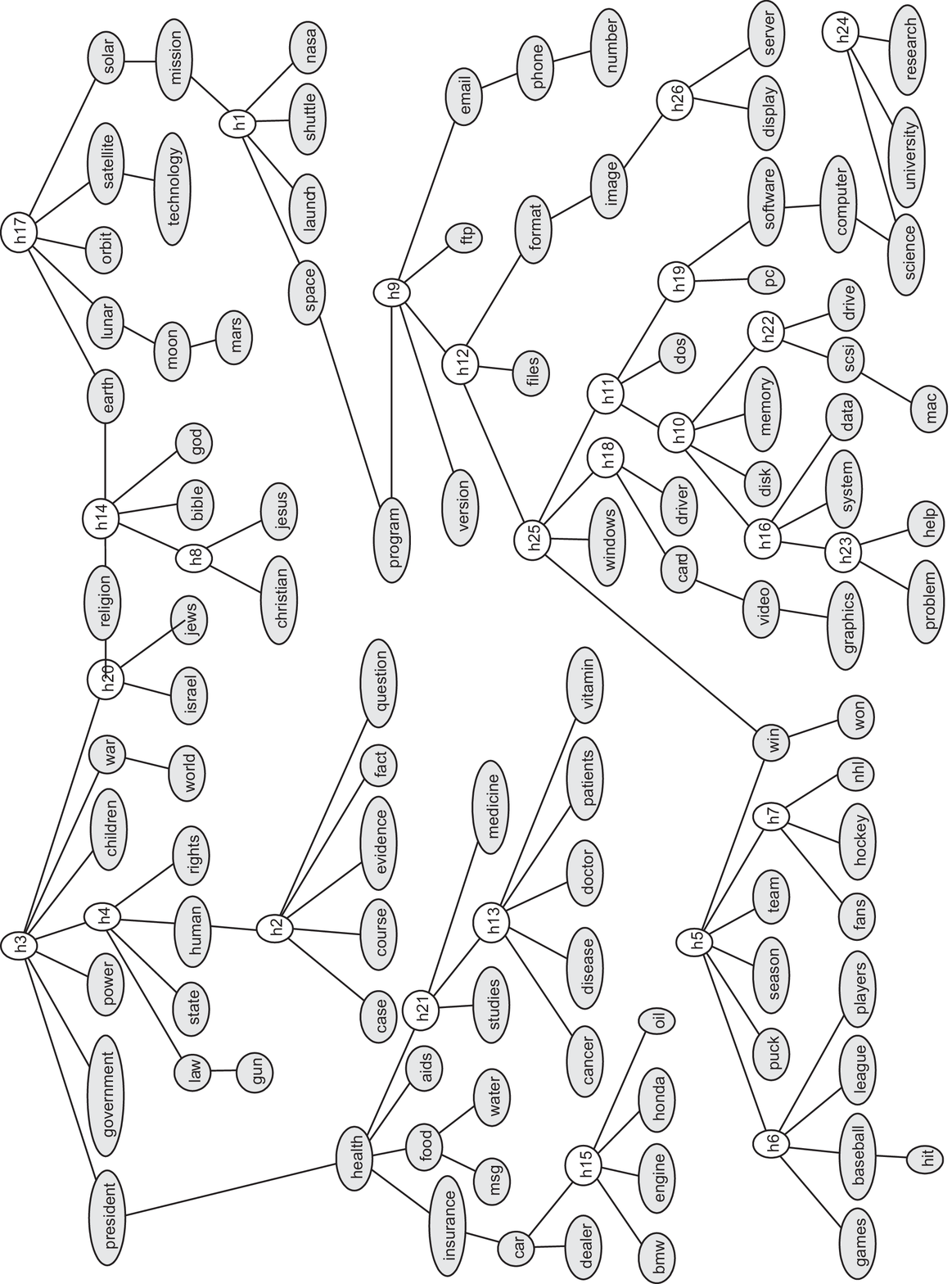}}
\caption{Tree structure learned from 20 newsgroup dataset using regCLRG.}
\label{fig:newsgroup_regCLRG_tree}
\end{figure}

Figure~\ref{fig:newsgroup_regCLRG_tree} shows the latent tree structure learned using regCLRG from the entire dataset.  Many hidden variables in the tree can be interpreted as topics - h5 as sports, h9 as computer technology, h13 as medical, etc.  Note that some words have multiple meanings and appear in different topics - e.g., {\tt program} can be used in the phrase ``space program'' as well as ``computer program'', and {\tt win} may indicate the windows operating system or winning in sports games.

\begin{table}[t]
\centering
 
\begin{tabular}{|c|c|c|c|c|c|c|c|}
\hline
  &\multirow{2}{*}{Log-Likelihood} & \multirow{2}{*}{BIC} & \multirow{2}{*}{Hidden} &  \multirow{2}{*}{Params} &  \multicolumn{3}{c|}{Time (s)} \\ \cline{6-8} 
  & & & & & Total & Structure & EM \\\hline
  CL & -238,713 & -239,677 & 0 & 199 & 8.9 & - & -\\\hline
  LCM & {\bf -223,096}& {\bf -230,925} & 1 & 1,615 & 8,835.9 & - &- \\\hline
BIN & -232,042 & -233,952 & 98 & 394& 3022.6 & - &- \\\hline
NJ& -230,575 & -232,257 & 74 & 347 & 1611.2 &3.3 &1608.2 \\\hline \hline
RG & -239,619 & -240,875 & 30 & 259 & 927.1& 30.8 & 896.4\\\hline 
CLNJ & -230,858 & -232,540 & 74 & 347 & 1479.6 & 2.7 & 1476.8\\\hline
CLRG & -231,279 & -232,738 & 51 & 301 & 1224.6 & 3.1 & 1224.6\\\hline
regCLNJ & -235,326 & -236,553 & 27 & 253 & 630.8 & 449.7 & 181.1\\\hline
regCLRG & -234,012 & -235,229 & 26 & 251 & 606.9 & 493.0 & 113.9\\\hline
\end{tabular}  
\caption{Comparison between various algorithms on the newsgroup set.}
\label{table:newsgroup_table_all}
\end{table}

\begin{table}[t]
\centering
{\scriptsize
\begin{tabular}{|c|c|c|c|c|c|c|c|c|c|}
\hline
& \multicolumn{2}{c|}{Train}  & \multicolumn{2}{c|}{Test}  & \multirow{2}{*}{Hidden} & \multirow{2}{*}{Params} & \multicolumn{3}{c|}{Time (s)}  \\  \cline{2-5} \cline{8-10}
& Log-Like & BIC & Log-Like & BIC & & & Total & Struct & EM\\ \hline
CL & -119,013 & -119,909 & -120,107 & -121,003 & 0 & 199 & 3.0 & - & -\\\hline
LCM & {\bf -112,746} & -117,288 & -116,884 & -120,949 & 1 & 1,009 & 3,197.7  & - & -\\\hline
BIN & -117,172 & -118,675& -117,957 & -119,460 & 78 &  334 & 1,331.3 & - & -\\\hline
NJ & -115,319 & {\bf -116,908} & {\bf -116,011} & -117,600 & 77 &  353 & 802.8 & 1.3 & 801.5\\\hline\hline
RG & -118,280 & -119,248 & -119,181 & -120,149 & 8&  215 & 137.6 & 7.6 & 130.0\\\hline
CLNJ & -115,372 & -116,987 & -116,036 & -117,652 & 80 &  359 & 648.0 & 1.5 & 646.5\\\hline
CLRG & -115,565 & -116,920 & -116,199 & {\bf -117,554} & 51 &  301 & 506.0& 1.7 & 504.3\\\hline
regCLNJ & -117,723 & -118,924 & -118,606 & -119,808 & 34 &  267 & 425.5& 251.3 & 174.2\\\hline
regCLRG & -116,980 & -118,119 & -117,652 & -118,791 & 27& 253 & 285.7 & 236.5 & 49.2\\\hline
\end{tabular} }
\caption{Comparison between various algorithms on the newsgroup dataset with a train/test split.}
\label{table:newsgroup_table_trte}
\end{table}

\section{Conclusion}
\label{sec:conclusion}

In this paper, we proposed algorithms to learn a latent tree model from the information distances of observed variables.  Our first algorithm, recursive grouping, identifies sibling and parent-child relationships and introduces hidden nodes recursively. Our second algorithm, CLGrouping, first learns the Chow-Liu tree among observed variables and then applies latent-tree-learning subroutines such as recursive grouping or neighbor joining locally to each internal node in the Chow-Liu tree and its neighbors. These algorithms are structurally consistent (and risk consistent as well in the case of Gaussian and discrete symmetric distributions), and have sample complexity logarithmic in the number of observed variables.  

Using simulations with synthetic datasets, we showed that RG performs well when the number of hidden variables is small, and that CLGrouping performs significantly better than other algorithms when there are many hidden variables in the latent tree.  Using both Gaussian and discrete real-world data, we compared the performances of our algorithms to other EM-based approaches and the neighbor-joining method, and our algorithm show superior results in both accuracy (measured by KL-divergence and graph distance) and computational efficiency.  In addition, we introduced regularized CLGrouping, which is useful in learning a latent tree approximation with a given number of hidden nodes.  The MATLAB implementation of our algorithms can be downloaded from the project webpage \url{http://people.csail.mit.edu/myungjin/latentTree.html}.


\subsubsection*{Acknowledgement}
The authors thank Prof.\ Sekhar Tatikonda (Yale University) for discussions and comments. This
work was supported in part by AFOSR under Grant FA9550-08-1-1080 and in part
by MURI under AFOSR Grant FA9550-06-1-0324. Vincent Tan is also supported by A*STAR, Singapore. 

\appendix

\section{Proofs} 
\subsection{Proof  of Lemma~\ref{th:gauss_family}: Sibling Grouping }
\label{app:th:gauss_family} We prove statement (i) in Lemma~\ref{th:gauss_family} using~\eqref{eqn:markov} in Proposition~\ref{fact:mst}. Statement  (ii) follows along similar lines and its proof is omitted for brevity.

{\em If :} From the additive property of information distances in~\eqref{eqn:markov}, if $i$ is a leaf node and $j$ is its parent, $d_{ik} = d_{ij}+ d_{jk}$ and thus $\Phi_{ijk} = d_{ij}$ for all $k \neq i,j$.

\begin{figure}[t]
\centerline{\includegraphics[width=0.8\linewidth]{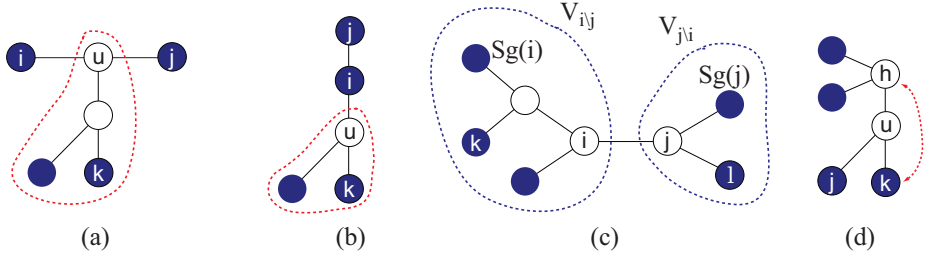}}
\caption{Shaded nodes indicate observed nodes and the rest indicate hidden nodes. (a),(b) Figures for Proof of Lemma~\ref{th:gauss_family}. Dashed red line represent the subtrees away from $i$ and $j$. (c) Figure for Proof of Lemma~\ref{lemma:surrogate}(i). (d) Figure for Proof of Lemma~\ref{lemma:surrogate}(iI) }
\label{fig:proofs}
\end{figure}

{\em Only If: } Now assume  that $\Phi_{ijk} = d_{ij}$ for all $k \in V\setminus \{i,j\}$. In order to prove that $i$ is a leaf node and $j$ is its parent,  assume to the contrary, that  $i$ and $j$ are not connected with an edge, then there exists a node $u \neq i,j$ on the path connecting $i$ and $j$.  If $u \in V$, then let $k = u$. Otherwise, let $k$ be an observed node in the subtree away from $i$ and $j$ (see Figure~\ref{fig:proofs}(a)), which exists since $T_p \in \calT_{\geq 3}$. By the additive property of information distances in~\eqref{eqn:markov} and the assumption that all distances are positive,
\begin{equation}
d_{ij} = d_{iu} +  d_{uj} > d_{iu} - d_{uj} = d_{ik}- d_{kj}  = \Phi_{ijk}
\end{equation}
which is a contradiction. If $i$ is not a leaf  node in $T_p$, then   there exist a node $u \neq i,j$ such that $(i,u) \in E_p$.  Let $k = u$ if $u \in V$, otherwise, let $k$ be an observed node in the subtree away from $i$ and $j$ (see Figure~\ref{fig:proofs}(b)). Then, 
\begin{equation}
\Phi_{ijk} = d_{ik}- d_{jk}  = -d_{ij} < d_{ij},
\end{equation}
which is again a contradiction. Therefore, $(i,j)\in E_p$ and   $i$ is a leaf node. \qed

\subsection{Proof of Theorem~\ref{th:correct_rg}: Correctness  and Computational Complexity of RG} \label{app:th:correct_rg}
The correctness of RG follows from the following observations: Firstly, at each iteration of RG, the sibling groups are identified correctly by Lemma~\ref{th:gauss_family}. Since the new parent node added to a partition which does not contain an observed parent corresponds  to a hidden node (in the original latent tree), a subforest of $T_p$ is recovered at each iteration.  Secondly, from Proposition~\ref{fact:mst}, for all $i,j$ in the active set $Y$,   the information distances $d_{ij}$ can be computed exactly with~\eqref{eq:dist_childparent_exact} and~\eqref{eq:dist_hiddenother_exact}.  Thirdly, since the information distances between nodes in the updated active set $Y$ are known, they can now be regarded as observed nodes. Subforests of $T_p$ are constructed at each iteration and when $|Y|\le 2$, the entire latent tree is recovered.

The computational complexity follows from the fact there are a maximum of  $O(m^3)$ differences $\Phi_{ijk}=d_{ik}-d_{jk}$ that we have to compute at each iteration of RG.  Furthermore, there are at most $\diam(T_p)$ subsets  in the coarsest partition (cf.\ step~\ref{item:partitions}) of $Y$ at the first iteration, and the number of subsets reduce at least by $2$ from one iteration to the next due to the assumption that $T_p \in \calT_{\geq 3}$.    This proves the claim that the computational complexity is upper bounded by $O(\diam(T_p)m^3)$.  \qed

\subsection{Proof  of Lemma~\ref{lemma:surrogate}: Properties of the MST}
\label{app:lemma:surrogate}


(i)  For an edge $(i,j)\in E_p$ such that $\Sg(i) \neq \Sg(j)$, let $V_{i\setminus j}\subset V$ and $V_{j\setminus i}\subset V$ denote observed nodes in the subtrees obtained by   the removal of edge $(i,j)$, where the former includes node $i$ and excludes node $j$ and vice versa (see Figure~\ref{fig:proofs}(c)).  Using part (ii) of the lemma and the fat that $\Sg(i) \neq \Sg(j)$, it can be shown that $\Sg(i) \in V_{i \setminus j}$ and $\Sg(j)\in V_{j \setminus i}$. Since $(i,j)$ lies on the unique path from $k$ to $l$  on $T_p$, for all observed nodes $k \in   V_{i \setminus j}, l \in   V_{j \setminus i}$, we have
\begin{equation}
d_{kl} = d_{ki} + d_{ij} + d_{jl} \ge d_{\Sg(i), i}+ d_{ij} + d_{\Sg(j), j} = d_{\Sg(i), \Sg(j)},
\end{equation}
where the inequality is  from   the definition of surrogacy and the final equality uses the fact that $\Sg(i)\neq \Sg(j)$. By using the property of the $\MST$ that $(\Sg(i), \Sg(j))$ is the shortest edge from $V_{i \setminus j}$ to $V_{j \setminus i}$, we have \eqref{eqn:surrogate:item1}.  

(ii) First assume that we have a tie-breaking rule consistent across all hidden nodes so that if $d_{uh} = d_{vh} = \min_{i \in V} d_{ih}$ and $d_{uh'} = d_{vh'} = \min_{i \in V} d_{ih'}$ then both $h$ and $h'$ choose the same surrogate node.  Let $j \in V$, $h \in \Sg^{-1}(j)$, and let $u$ be a node on the path connecting $h$ and $j$ (see Figure~\ref{fig:proofs}(d)).  Assume that $\Sg(u) = k \neq j$.  If $d_{uj} > d_{uk}$, then
\begin{equation}
d_{hj} = d_{hu} + d_{uj} > d_{hu} + d_{uk} = d_{hk},
\end{equation}
which is a contradiction since $j = \Sg(h)$.  If $d_{uj} = d_{uk}$, then $d_{hj} = d_{hk}$, which is again a contradiction to the consistent tie-breaking rule.  Thus, the surrogate node of $u$ is $j$.

(iii) First we claim that 
\begin{equation}
|\Sg^{-1}(i)|\le \Delta(T_p)^{\frac{u}{l} \delta(T_p;V)}. \label{eqn:claim_surrogate}
\end{equation}
To prove this claim, let $\gamma$ be the longest (worst-case) graph distance of any hidden node $h\in H$ from its surrogate, i.e., 
\begin{equation}
\label{eqn:gamma_def} 
\gamma:= \max_{h\in H}|\Path(h,\Sg(h);T_p)|.
\end{equation} 
 From the degree bound, for each $i \in V$, there are at most $\Delta(T_p)^{\gamma}$ hidden nodes that are within the graph distance of $\gamma$,\footnote{\scriptsize The maximum size of the inverse surrogate set in \eqref{eqn:gamma_def} is attained by a $\Delta(T_p)$-ary complete tree.} so 
\begin{equation}
|\Sg^{-1}(i)|\le \Delta(T_p)^{\gamma} \label{eqn:size_sginv}
\end{equation}
for all $i\in V$.  Let $d^*:= \max_{h\in H}  d_{h,\Sg(h)}$ be the longest (worst-case) information distance between a hidden node and its surrogate.  
From the bounds on the information distances, $l\gamma\le d^*$. In addition, for each $h \in H$, let $z(h) := \argmin_{j \in V} | \Path((h,j); T_p) |$ be the observed node that is closest to $h$ in graph distance.  Then, by definition of the effective depth, $d_{h, \Sg(h)} \le d_{h, z(h)} \le u \delta$ for all $h \in H$, and we have $d^* \le u \delta$.  Since $l\gamma\le d^*  \le u \delta$,  we also have 
\begin{equation}\label{eqn:gamma}\gamma\le u\delta/l.\end{equation} Combining this result with \eqref{eqn:size_sginv} establishes the claim in \eqref{eqn:claim_surrogate}. Now consider  
\begin{equation}
\Delta(\MST(V;\bD)) ~\stackrel{(a)}{\le}~ \Delta(T_p)\max_{i\in V}|\Sg^{-1}(i)| ~\stackrel{(b)}{\le}~ \Delta(T_p)^{1+ \frac{u}{l} \, \Depth(T_p;V)}
\end{equation}
where $(a)$ is a result of the application of  \eqref{eqn:surrogate:item1} and $(b)$ results from \eqref{eqn:claim_surrogate}. This completes the proof of the claim in \eqref{eqn:surrogate:item2} in Lemma~\ref{lemma:surrogate}. \qed


\subsection{Proof  of Theorem~\ref{th:CLBlind}: Correctness and Computational Complexity of CLBlind }
\label{app:th:CLBlind}


It suffices to show that the Chow-Liu tree   $\MST(V;\bd)$ is a transformation of the true latent tree $T_p$ (with parameters such that $p\in\calP(\calT_{\mathrm{blind}})$) as follows: contract the edge connecting each hidden variable $h$ with its surrogate node $\Sg(h)$ (one of its children and a leaf by assumption). Note that the blind transformation on the $\MST$ is merely the inverse mapping of the above. From \eqref{eqn:surrogate:item1}, all the children of a hidden node $h$, except its surrogate $\Sg(h)$, are neighbors of its surrogate node $\Sg(h)$ in $\MST(V;\bd)$. Moreover, these children of $h$ which are not surrogates of any hidden nodes are leaf nodes in the MST.  Similarly for two hidden nodes $h_1,h_2\in H$ such that $(h_1, h_2)\in E_p$, $(\Sg(h_1), \Sg(h_2))\in \MST(V;\bd)$ from Lemma~\ref{lemma:surrogate}(i).  Hence, CLBlind outputs the correct tree structure $T_p$. The computational complexity follows from the fact that the blind transformation is linear in the number of internal nodes, which is less than the number of observed nodes, and that learning the Chow-Liu tree takes $O(m^2 \log m)$ operations.\qed

\begin{figure}[t]
\centerline{\includegraphics[width=\linewidth]{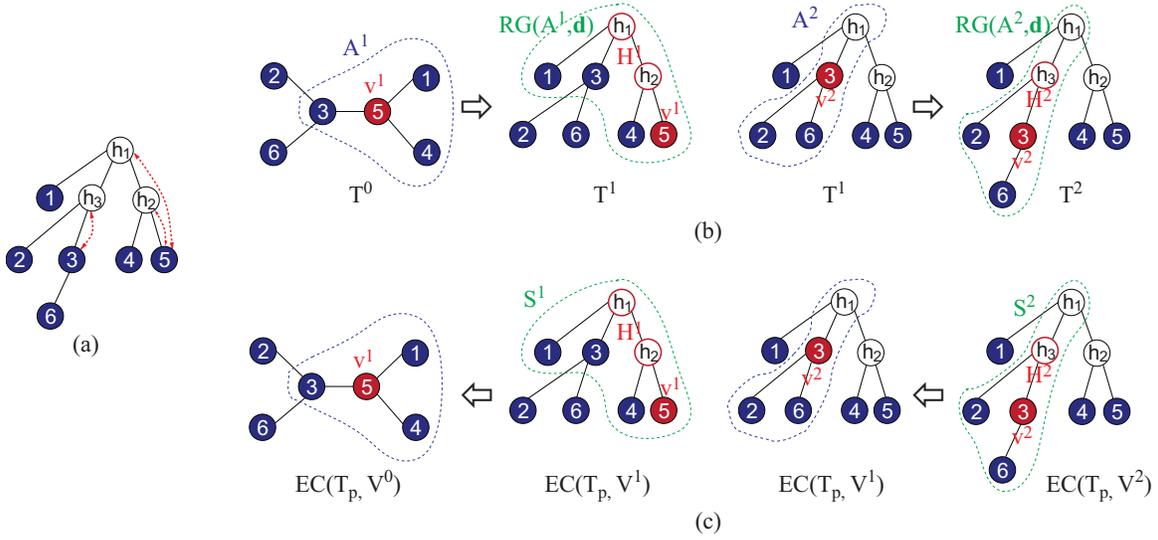}}
\caption{Figure for Proof of Theorem~\ref{th:CLGrouping}.  (a) Original latent tree. (b) Illustration of CLGrouping. (c) Illustration of the trees constructed using edge contractions. }
\label{fig:proof_clg}
\end{figure}

\subsection{Proof  of Theorem~\ref{th:CLGrouping}: Correctness and Computational Complexity of CLRG }
\label{app:th:CLGrouping}

We first define some new notations.

{\em Notation: } Let $\Ic:= V\setminus \Leaf(\MST(V;\bd))$ be the set of internal nodes. Let  $v^{r} \in \Ic$ be the  internal node visited  at iteration $r$, and let $H^r$ be all hidden nodes in the inverse surrogate set $\Sg^{-1}(v^r)$, i.e., $H^r = \Sg^{-1}(v^r) \setminus \{v^r\}$. Let $A^r:=\nbd[v^r;T^{r-1}]$, and hence $A^r$ is the node set input to the recursive grouping routine at iteration $r$, and let $\RG(A^r, \bd)$ be the output latent tree learned by recursive grouping.  Define  $T^{r}$ as the tree output  at the end of  $r$ iterations of CLGrouping.  Let $V^{r} := \{v^{r+1},v^{r+2},\ldots, v^{|\Ic|}\}$ be the set of internal nodes that have not yet been visited by CLGrouping at the end of $r$ iterations.  Let $\EC(T_p,V^{r})$ be the tree constructed using edge contractions as follows:  in the latent tree $T_p$, we contract edges corresponding to each node $u \in V^{r}$ and all hidden nodes in its inverse surrogate set $\Sg^{-1}(u)$.  Let $S^r$ be a subtree of $\EC(T_p,V^{r})$ spanning $v^r$, $H^r$ and their neighbors. 


For example, in Figure~\ref{fig:proof_clg}, the original latent tree $T_p$ is shown in Figure~\ref{fig:proof_clg}(a), and $T^0$, $T^1$, $T^2$ are shown in Figure~\ref{fig:proof_clg}(b). The set of internal nodes is $\Ic = \{3,5\}$. In the first iteration, $v^1=5$, $A^1=\{1,3,4,5\}$ and $H^1 = \{h_1, h_2\}$. In the second iteration, $v^2=3$,  $A^2=\{2,3,6,h_1\}$ and $H^1 = \{h_3\}$. $V^0 = \{3,5\}$, $V^1 = \{3\}$, and $V^2 = \emptyset$, and in Figure~\ref{fig:proof_clg}(c), we show $\EC(T_p,V^{0})$, $\EC(T_p,V^{1})$, and $\EC(T_p,V^{2})$. In $\EC(T_p,V^{1})$, $S^1$ is the subtree spanning $5, h_1, h_2$ and their neighbors, i.e., $\{1,3,4,5,h_1,h_2\}$. In $\EC(T_p,V^{2})$, $S^2$ is the subtree spanning $3, h_3$ and their neighbors, i.e., $\{2,3,6,h_1,h_3\}$. Note that $T^0 = \EC(T_p,V^{0})$, $T^1 = \EC(T_p,V^{1})$, and $T^2 = \EC(T_p,V^{2})$; we show below that this holds for all CLGrouping iterations in general.


\vspace{1em}

We prove the theorem by induction on the iterations $r=1,\ldots, |\Ic|$ of the CLGrouping algorithm.  

{\em Induction Hypothesis:} At the end of $k$ iterations of CLGrouping, the tree obtained is \begin{equation}\label{eqn:induction_claim} T^{k} = 
\EC(T_p,V^{k}),\qquad\forall\,  k =0,1,\ldots,  |\Ic|.\end{equation}  
\noindent In words, the latent tree after $k$ iterations of CLGrouping can be constructed by contracting each surrogate node in $T_p$ that has not been visited by CLGrouping with its inverse surrogate set. Note that $V^{|\Ic|}= \emptyset$ and  $\EC(T_p,V^{|\Ic|})$ is equivalent to the original latent tree $T_p$.  Thus, if the above induction in \eqref{eqn:induction_claim} holds,  then the output of CLGrouping $T^{|\Ic|}$  is the original latent tree. 

{\em Base Step $r=0$:} The claim in \eqref{eqn:induction_claim} holds since $V^{0}=\Ic$ and the input to the CLGrouping procedure is the Chow-Liu tree $\MST(V; \bD)$, which is obtained by contracting all surrogate nodes and their inverse surrogate sets (see Section \ref{sec:cl_latent}).

{\em Induction Step:} Assume \eqref{eqn:induction_claim} is true for $k=1,\ldots,r-1$. Now consider $k=r$.

We first compare the two latent trees $\EC(T_p,V^{r})$ and $\EC(T_p,V^{r-1})$.   By the definition of $\EC$, if we contract edges with $v^r$ and the hidden nodes in its inverse surrogate set $H^r$ on the tree $\EC(T_p,V^{r})$, then we obtain $\EC(T_p,V^{r-1})$, which is equivalent to $T^{r-1}$ by the induction assumption.  Note that as shown in Figure~\ref{fig:proof_clg}, this transformation is local to the subtree $S^r$:  contracting $v^r$ with $H^r$  on $\EC(T_p,V^{r})$ transforms $S^r$ into a star graph with $v^r$ at its center and the hidden nodes $H^r$ removed (contracted with $v^r$). 

Recall that the CLGrouping procedure replaces the induced subtree of $A^r$  in $T^{r-1}$ (which is precisely the star graph mentioned above by the induction hypothesis) with $\RG(A^r,\bd)$ to obtain $T^r$.  Thus, to prove that $T^r = \EC(T_p,V^{r})$, we only need to show that RG reverses the edge-contraction operations on $v^r$ and $H^r$, that is, the subtree $S^r=\RG(A^r,\bd)$.  We first show that  $S^r \in \calT_{\geq 3}$, i.e., it is identifiable (minimal) when $A^r$ is the set of visible nodes. This is because an edge contraction operation does not decrease the degree of any existing nodes.  Since $T_p\in \calT_{\geq 3}$, all hidden nodes in $\EC(T_p,V^{r})$ have degrees equal to or greater than $3$, and since we are including all neighbors of $H^r$ in the subtree $S^r$, we have $S^r \in \calT_{\geq 3}$. By Theorem~\ref{th:correct_rg}, RG reconstructs all latent trees in $\calT_{\geq 3}$ and hence, $S^r=\RG(A^r,\bd)$.

\vspace{1em}

The computational complexity  follows from the corresponding result in  recursive grouping. The Chow-Liu tree can be constructed with $O(m^2 \log m)$ complexity. The recursive grouping procedure has complexity $\max_r |A^r|^3$ and $\max_r |A^r|\leq \Delta(\MST(V;\hd))$. \qed

\subsection{Proof of Theorem~\ref{thm:rg_sample}: Consistency and Sample Complexity of Relaxed RG}\label{proof:rg_sample}

(i) Structural consistency follows from Theorem~\ref{th:correct_rg} and the fact that the ML estimates of information distances $\hd_{ij}$ approach $d_{ij}$  (in probability) for all $i,j \in V$ as the number of samples tends to infinity.

Risk consistency for Gaussian and symmetric discrete distributions follows from structural consistency. If the structure is correctly recovered, we can use the equations in \eqref{eq:dist_childparent_exact} and \eqref{eq:dist_hiddenother_exact} to infer the information distances. Since the distances are in one-to-one correspondence to the correlation coefficients and the crossover probability for Gaussian and symmetric discrete distribution respectively, the parameters are also consistent. This implies that the KL-divergence between $p$ and $\hp^n$ tends to zero (in probability) as the number of samples  $n$ tends to infinity. This completes the proof. 

\vspace{1em}

(ii)  The theorem follows by using the assumption that the effective  depth $\delta=\delta(T_p; V)$ is constant. Recall that $\tau>0$ is the threshold used in relaxed RG (see~\eqref{eqn:calKij} in Section~\ref{subsec:rg_estimate}).  
Let the set of triples $(i,j,k)$ whose pairwise information distances are less than $\tau$ apart be  $\calJ$, i.e., $(i,j,k)\in\calJ$ if and only if $\max\{ d_{ij}, d_{jk}, d_{ki}\}<\tau$.  Since we assume that the true information distances are uniformly bounded, there exist $\tau > 0$ and some sufficiently small $\lambda>0$ so that if $|\hPhi_{ijk}-\Phi_{ijk}|\le \lambda$ for all $(i,j,k)\in\calJ$, then RG recovers the correct latent structure.


Define the error event $\calE_{ijk} :=\{ |\hPhi_{ijk}-\Phi_{ijk}|>\lambda\}$. We   note that the probability of the event  $\calE_{ijk}$ decays exponentially fast, i.e., there exists $J_{ijk}>0$ such that  for all $n\in\bN$,
\begin{equation} \label{eqn:Eijk}
\Pr(\calE_{ijk}) \le  \exp(-n J_{ijk} ).
\end{equation}
The proof of \eqref{eqn:Eijk} follows readily for   Chernoff bounds \citep{Hoe58} and is omitted. The error probability associated to structure learning can be bounded as follows:
\begin{align}
\Pr\left(h(\hT^n) \ne T_p \right)& ~\stackrel{(a)}{\le}~  \Pr\left(\bigcup_{(i,j,k)\in\calJ} \calE_{ijk} \right)  
 ~\stackrel{(b)}{\le}~ \sum_{(i,j,k)\in \calJ}\Pr(\calE_{ijk}) \\
&~\le~ m^3 \max_{(i,j,k)\in \calJ}\Pr(\calE_{ijk}) 
~ \stackrel{(c)}{\le}~ \exp(3\log m) \exp\left[-n \min_{(i,j,k)\in \calJ} J_{ijk} \right],  \label{eqn:Jijk}
\end{align}
where $(a)$ follows from the fact that if the event $\{h(\hT^n)\ne T_p\}$ occurs, then there is at least one sibling or parent-child relationship that is incorrect, which corresponds to the union of the events $\calE_{ijk}$, i.e., there exists a triple $(i,j,k)\in \calJ$ is such that $\hPhi_{ijk}$ differs from $\Phi_{ijk}$ by more than $\lambda$. Inequality $(b)$ follows from the union   bound and $(c)$ follows from~\eqref{eqn:Eijk}.  

Because the information distances are uniformly bounded, there also exists  a constant $J_{\min}>0$ (independent of $m$) such that     $\min_{(i,j,k)\in \calJ} J_{ijk}\ge J_{\min}$ for all $m\in\bN$. Hence for every $\eta>0$, if the number of samples satisfies $n>3(\log (m/\sqrt[3]{\eta}))/J_{\min}$, the error probability is bounded above by $\eta$.  Let $C:=3/J_{\min}$ to complete the proof of the sample complexity result in~\eqref{eqn:sample_complex_rg}. The proof for the logarithmic sample complexity of distribution reconstruction for Gaussian and symmetric discrete  models follows from the logarithmic sample complexity result for structure learning  and the fact that the information distances are in a one-to-one correspondence with the correlation coefficients (for Gaussian models) or crossover probabilities (for symmetric discrete models).


\subsection{Proof of Theorem~\ref{th:cl_sample}: Consistency and Sample Complexity of Relaxed CLRG }
\label{proof:cl_sample}

(i) Structural consistency of CLGrouping follows from structural consistency of RG (or NJ) and the consistency of the Chow-Liu algorithm.  Risk consistency of CLGrouping for Gaussian or symmetric distributions follows from the structural consistency, and the proof is similar to the proof of Theorem~\ref{th:rg_sample}(i).

\vspace{1em}

(ii) The input to the CLGrouping procedure $\widehat{T}_{\ML}$ is the Chow-Liu tree and has $O(\log m)$ sample complexity \cite{tan10}, where $m$ is the size of the tree. From Theorem~\ref{th:rg_sample}, the recursive grouping procedure has $O(\log m)$ sample complexity (for appropriately chosen thresholds) when the input information distances are uniformly bounded. In any iteration of the CLGrouping, the information distances satisfy $d_{ij}\leq \gamma u$, where  $\gamma$, defined in \eqref{eqn:gamma_def}, is the worst-case graph distance of any hidden node  from its surrogate. Since $\gamma$ satisfies \eqref{eqn:gamma}, $d_{ij}\le u^2\delta/l$. If the effective depth $\delta=O(1)$ (as assumed), the distances  $d_{ij}=O(1)$ and  the sample complexity is $O(\log m)$. \qed

\bibliography{latentTree_ref}

\end{document}